\documentclass[runningheads]{llncs}
\usepackage{graphicx}
\usepackage{comment}
\usepackage{amsmath,amsfonts,amssymb,mathrsfs}
\usepackage{color}
\usepackage[width=122mm,left=12mm,paperwidth=146mm,height=193mm,top=12mm,paperheight=217mm]{geometry} 
\usepackage{pdflscape}

\usepackage[svgnames,table]{xcolor}
\usepackage{epsfig}
\usepackage{makecell}

\usepackage{graphicx}
\usepackage[ruled]{algorithm2e}
\usepackage{amsmath,amssymb}
\usepackage{epstopdf}
\usepackage{tabularx}
\usepackage{eqnarray}
\usepackage{array,booktabs}
\usepackage{enumitem}
\usepackage[space]{grffile} 
\usepackage{soul} 
\usepackage{float}
\usepackage{tabularx,ragged2e} 
\usepackage{bbm}
\usepackage{pifont} 
\usepackage{cancel}
\usepackage{wrapfig}
\usepackage{color, colortbl}

\makeatletter

\renewcommand{\paragraph}{%
  \@startsection{paragraph}{4}%
  {\z@}{0.3ex \@plus 1ex \@minus .2ex}{-1em}
  {\normalfont\normalsize\bfseries}%
}
\makeatother

\definecolor{myGray}{rgb}{0.6,0.6,0.6}
\definecolor{myBlue}{rgb}{0.1,0.1,0.8}

\def\onedot{\ifx\@let@token.\else.\null\fi\xspace}

\def\eg{\emph{e.g}\onedot} 
\def\ie{\emph{i.e}\onedot}

\def\etal{\emph{et al}\onedot}

\definecolor{pcGray}{rgb}{0.5,0.5,0.5}

\definecolor{pccGray}{rgb}{0.3,0.3,0.3}

\newcommand{\fig}{Fig.~}
\newcommand{\eq}{Eq.\,}
\newcommand{\sect}{Section~}
\newcommand{\tab}{Table~}

\newcommand{\hangBoxC}[1]{%
  \begin{minipage}[c]{\textwidth}\begin{tabbing} 
  ~\\[-\baselineskip] 
  #1 
  \end{tabbing}
  \end{minipage}}

\newcommand\minput[1]{%
  \input{#1}%
  \ifhmode\ifnum\lastnodetype=11 \unskip\fi\fi}

\usepackage{listings}
\usepackage{color}

\ifx \@etal \@empty \newcommand{\etal}{et al.} \fi
\ifx \@eg \@empty \newcommand{\eg}{e.g.,~} \fi
\ifx \@ie \@empty \newcommand{\ie}{i.e.,~} \fi
\ifx \@etc \@empty  \fi

\newcommand{\ba}{\boldsymbol{a}}

\newcommand{\bg}{\boldsymbol{g}}

\newcommand{\bq}{{\boldsymbol{q}}}

\newcommand{\bx}{{\boldsymbol{x}}}
\newcommand{\by}{{\boldsymbol{y}}}

\newcommand{\bI}{{\boldsymbol{I}}}

\newcommand{\btheta}{{\boldsymbol{\theta}}}

\newcommand{\tablefont}{\small}

\newcommand{\gradf}{{\nabla_\bx f}}

\begin{document}
\pagestyle{headings}
\mainmatter
\def\ECCVSubNumber{1165}

\title{Learning What Makes a Difference\\from Counterfactual Examples and\\Gradient Supervision}

\titlerunning{Learning What Makes a Difference}
\author{Damien Teney, Ehsan Abbasnedjad, Anton van den Hengel}
\authorrunning{Teney et al.}
\institute{Australian Institute for Machine Learning\\
University of Adelaide\\
North Terrace, SA 5005 Adelaide, Australia
}

\maketitle

\begin{abstract}
One of the primary challenges limiting the applicability of deep learning is its susceptibility to learning spurious correlations rather than the underlying mechanisms of the task of interest. The resulting failure to generalise cannot be addressed by simply using more data from the same distribution. We propose an auxiliary training objective that improves the generalization capabilities of neural networks by leveraging an overlooked supervisory signal found in existing datasets. We use pairs of minimally-different examples with different labels, a.k.a counterfactual or contrasting examples, which provide a signal indicative of the underlying causal structure of the task. We show that such pairs can be identified in a number of existing datasets in computer vision (visual question answering, multi-label image classification) and natural language processing (sentiment analysis, natural language inference). The new training objective orients the gradient of a model's decision function with pairs of counterfactual examples. Models trained with this technique demonstrate improved performance on out-of-distribution test sets.
\end{abstract}

\section{Introduction}
\label{intro}

\begin{figure}[t!]
	\centering
	\includegraphics[width=1\linewidth]{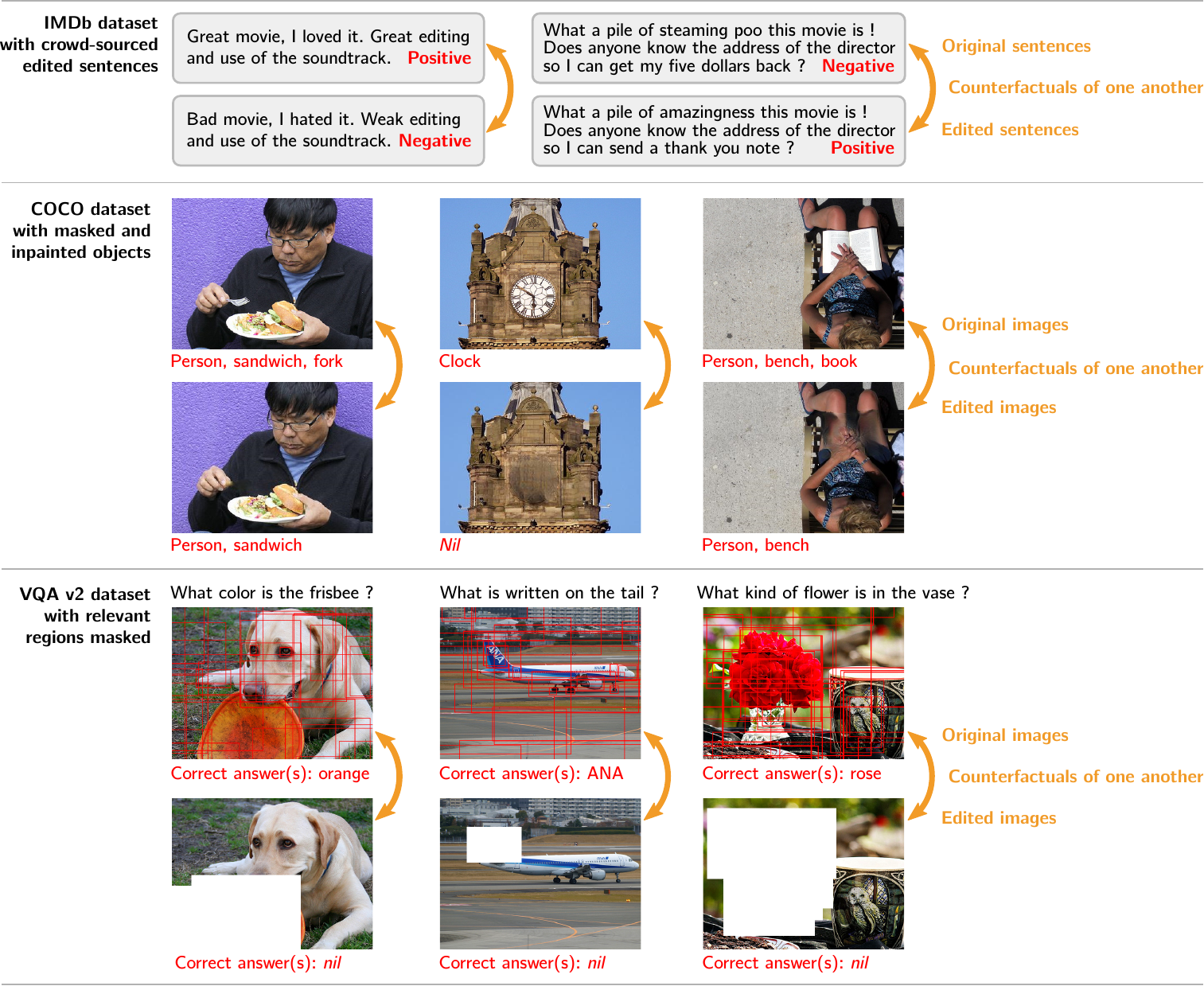}
	\vspace{-15pt}
	\caption{
	The traditional supervised learning process treats each training example individually.
	We use counterfactual \emph{relations} between examples as an additional supervisory signal.
	In some datasets, the relations are provided explicitly, as in sentiment analysis (top) with sentences edited by annotators to flip their label between positive and negative.
	In other datasets (COCO, middle, and VQA v2, bottom), we show that counterfactual examples can be generated from existing annotations, by masking relevant regions.
	Our method then leverages the \emph{relations} between the original and edited examples, which proves superior to simple data augmentation.
	\label{figTeaser}}
	\vspace{-8pt}
\end{figure}

Most of today's machine learning methods rely on the assumption that the training and testing data are drawn from a same distribution~\cite{vapnik1999overview}.
One implication is that models are susceptible to poor real-world performance when the test data differs from what is observed during training.
This limited capability to generalise partly arises because supervised training essentially amounts to identifying correlations between given examples and their labels. 
However, correlations can be spurious, in the sense that they may reflect dataset-specific biases or sampling artifacts, rather than intrinsic properties of the task of interest~\cite{mitchell1980need,torralba2011unbiased}.
When spurious correlations do not hold in the test data, the model's predictive performance suffers and its output becomes unreliable and unpredictable.
For example, an image recognition system may rely on common co-occurrences of objects, such as people together with a dining table, rather visual evidence for each recognized object.
This system could then hallucinate people when a table is observed (\fig\ref{figCoco}).

A model capable of generalization and extrapolation beyond its training distribution should ideally capture the causal mechanisms at play behind the data.
Acquiring additional training examples from the same distribution cannot help in this process~\cite{pearl2000causality}.
Rather, we need either to inject strong prior assumptions in the model, such as inductive biases encoded in the architecture of a neural network, or a different type of training information.
Ad hoc methods such as data augmentation and domain randomization fall in the former category, and they only defer the limits of the system by hand-designed rules.

In this paper, we show that many existing datasets contain an overlooked signal that is informative about their causal data-generating process.
This information is present in the form of groupings of training examples, and it is often discarded by the shuffling of points occurring during stochastic training.
We show that this information can be used to learn a model that is more faithful to the causal model behind the data.
This training signal is fundamentally different and complementary to the labels of individual points. 
We use pairs of minimally-dissimilar, differently-labeled training examples, which we interpret as counterfactuals of one another.
In some datasets, such pairs are provided explicitly~\cite{baradel2019cophy,camburu2018snli,park2019robust,park2019viewpoint}. In others, they can be identified from existing annotations~\cite{das2016attention,goyal2017making,lin2014microsoft,suhr2017corpus,suhr2018corpus,vo2019composing}.

The intuition for our approach is that relations between pairs of counterfactual examples indicate what changes in the input space map to changes in the space of labels.
In a classification setting, this serves to constrain the geometry of a model's decision boundary between classes.
Loosely speaking, we complement the traditional ``curve fitting'' to individual training points of standard supervised learning, with ``aligning the curve'' with pairs of counterfactual training points.

We describe a novel training objective (gradient supervision) and its implementation on various architectures of neural networks.
The vector difference in input space between pairs of counterfactual examples serves to supervise the orientation of the gradient of the network.
We demonstrate the benefits of the method on four tasks in computer vision and natural language processing (NLP) that are notoriously prone to poor generalization due to dataset biases: visual question answering (VQA), multi-label image classification, sentiment analysis, and natural language inference.
We use annotations from existing datasets that are usually disregarded, and we demonstrate significant improvements in generalization to out-of-distribution test sets for all tasks.

In summary, the contributions of this paper are as follows.
\setlist{nolistsep,leftmargin=*}
\begin{enumerate}[noitemsep]
	\item We propose to use relations between training examples as additional information in the supervised training of neural networks (\sect\ref{sectMotivation}). We show that they provide a fundamentally different and complementary training signal to the fitting of individual examples, and explain how they improve generalization (\sect\ref{sectTheory}).
	\item We describe a novel training objective (gradient supervision) to use this information and its implementation on multiple architectures of neural networks (\sect\ref{sectApplications}).
	\item We demonstrate that the required annotations are present in a number of existing datasets in computer vision and NLP, although they are usually discarded. We show that our technique brings improvements in out-of-distribution generalization on VQA, multi-label image classification, sentiment analysis, and natural language inference.
\end{enumerate}

\section{Related work}

This work proposes a new training objective that improves the generalization capabilities of models trained with supervision. This touches a number of core concepts in machine learning.

The predictive performance of machine learning models rests on the fundamental assumption of statistical similarity of the distributions of training and test data.
There is a growing interest for evaluating and addressing the limits of this assumption.
Evaluation on \textbf{out-of-distribution data} is increasingly common in computer vision~\cite{agrawal2018don,barbu2019objectnet,guo2019quantifying} and NLP~\cite{jia2017adversarial,zellers2018swag}.
These evaluations have shown that some of the best models can be right for the wrong reasons~\cite{agrawal2018don,feng2019misleading,goyal2017making,torralba2011unbiased,vapnik2019rethinking}. 
This happens when they rely on dataset-specific biases and artifacts rather than intrinsic properties of the task of interest.
When these biases do not hold in the test data, the predictive performance of the models can drop dramatically~\cite{agrawal2018don,barbu2019objectnet}.

When poor generalization is viewed as a deficiency of the training data, it is often referred to as \textbf{dataset biases}. They correspond to correlations between inputs and labels in a dataset that can be exploited by a model to exhibit strong performance on a test set containing these same biases, without actually solving the task of interest.
Several popular datasets used in vision-and-language~\cite{goyal2016balanced} and NLP~\cite{zellers2018swag} have been shown to exhibit strong biases, leading to an inflated sense of progress on these tasks.

Recent works have discussed generalization from a \textbf{causal perspective}~\cite{arjovsky2019invariant,heinze2017conditional,peters2016causal,rojas2018invariant}. This sheds light on the possible avenues for addressing the issue.
In order to generalize perfectly, a model should ideally capture the real-world causal mechanisms at play behind the data. The limits of identifiability of causal models from observational data have been well studied~\cite{pearl2000causality}.
In particular, additional data from a single biased training distribution can not solve the problem. The alternative options are to use strong assumptions (\eg inductive biases, engineered architectures, hand-designed data augmentations), or additional data, collected in controlled conditions and/or of a different type than labeled examples.
This work uses the latter option, using pairings of training examples that represent counterfactuals of one another. Recent works that follow this line include the principle of invariant risk minimization (IRM~\cite{arjovsky2019invariant}).
IRM uses multiple training environments, \ie non-IID training distributions, to discover generalizable invariances in the data.
Teney \etal~\cite{teney2020unshuffling} showed that existing datasets could be automatically partitioned to create these environments, and demonstrated improvements in generalization for the task of visual question answering (VQA).

Generalization is also related to the wide area of \textbf{domain adaptation}~\cite{ganin2016domain}.
Our objective in this paper is not to adapt to a particular new domain, but rather to learn a model that generalizes more broadly by using annotations indicative of the causal mechanisms of the task of interest.
In domain adaptation, the idea of finding a data representation that is invariant across domains is limiting, because the true causal factors that our model should rely on may differ in their distribution across training domains.
We refer the reader to \cite{arjovsky2019invariant} for a formal discussion of these issues.

The growing popularity of high-level tasks in \textbf{vision-and-language}~\cite{anderson2018vision,antol2015vqa,das2017dialog} has brought the issue of dataset biases to the forefront. In VQA, language biases cause models to be overly reliant on the presence of particular words in a question. Improving the data collection process can help~\cite{goyal2016balanced,zhang2015balanced} but it only addresses precisely identified biases and confounders.
Controlled evaluations for VQA now include out-of-distribution test sets~\cite{agrawal2018don,teney2016zsvqa}.
Several models and training methods~\cite{cadene2019rubi,clark2019don,teney2019incorporating,grand2019adversarial,guo2019quantifying,mahabadi2019simple,ramakrishnan2018overcoming} have been proposed with significant improvements. They all use strong prior knowledge about the task and/or additional annotations (question types) to improve generalization.
Some methods also supervise the model's attention~\cite{liu2017attention,qiao2018exploring,selvaraju2019taking} with ground truth human attention maps~\cite{das2016attention}.
All of these methods are specific to VQA or to captioning~\cite{hendricks2018women,liu2017attention} whereas we describe a much more general approach.

Evaluating generalization overlaps with the growing interest in \textbf{adversarial examples} for evaluation~\cite{bartolo2020beat,chen2019codah,iyyer2018adversarial,jia2017adversarial,miyato2016adversarial,nie2019adversarial,wallace2018trick}. The term has been used to refer both to examples purposefully generated to fool existing models \cite{moosavi2017universal,nguyen2015deep},
but also to hard natural examples that current models struggle with~\cite{bartolo2020beat,chen2019codah,wallace2018trick,zellers2019hellaswag}.
Our method is most related to the use of these examples for \textbf{adversarial training}. Existing methods focus mostly on the generation of these examples then mix them with the original data in a form of data augmentation~\cite{kaushik2019learning,shafahi2019adversarial,xie2019adversarial}. We argue that this shuffling of examples destroys valuable information. In many datasets, we demonstrate that relations between training points contain valuable information.
The above methods also aim at improving robustness to targeted adversarial attacks, which often use inputs outside the manifold of natural data.
Most of them rely on prior knowledge and unsupervised regularizers \cite{lirobust,jakubovitz2018improving,woods2019adversarial} whereas we seek to exploit additional supervision to improve generalization on natural data.

\section{Proposed approach}

We start with an intuitive motivation for our approach, then describe its technical realization.
In \sect\ref{sectTheory}, we analyze more formally how it can improve generalization.
In \sect\ref{sectApplications}, we demonstrate its application to a range of tasks.

\subsection{Motivation}
\label{sectMotivation}

Training a machine learning model amounts to fitting a function $f(\cdot)$ to a set of labeled points.
We consider a binary classification task, in which the model is a neural network $f$ of parameters $\btheta$ such that $f_\btheta : \mathbb{R}^d \rightarrow \{0,1\}$, and a set of training points%
\footnote{By \textit{input space}, we refer to a space of feature representations of the input, \ie vector representations ($\bx$) obtained with a pretrained CNN or text encoder.}
 $\mathcal{T}=\{(\bx_i,y_i)\}_i$, with $\bx_i \in \mathbb{R}^d$ and $y_i \in [0,1]$.
By training the model, we typically optimize $\btheta$ such that the output of the network
$\tilde{y}_i=f_\btheta(\bx_i)$ minimizes a loss
$\mathcal{L}_\mathrm{Main}(\tilde{y}_i, y_i)$
on the training points.
However, this does not specify the behaviour of $f$ between these points, and the decision boundary could take an arbitrary shape (\fig\ref{figPrinciple}).
The typical practice is to restrain the space of functions $\mathcal{F} \supset f$ (\eg a particular architecture of neural networks) and of parameters $\Theta \supset \btheta$ (\eg with regularizers~\cite{jakubovitz2018improving,lirobust,woods2019adversarial}).
The capability of the model to interpolate and extrapolate beyond the training points depends on these choices.

Our motivating intuition is that many datasets contain information that is indicative of the shape of an ideal $f$ (in the sense of being faithful to the data-generating process, see \sect\ref{sectTheory}) between training points.
In particular, we are interested in pairs of training examples that are \textbf{counterfactuals} of one another.
Given a labeled example $(\bx_1,y_1)$, we define its counterfactuals as examples such as $(\bx_2,y_2)$ that represents an alternative premise $\bx_2$ (``counter to the facts'') that lead to different outcome $y_2$.
These points represent ``minimal changes'' ($||\bx_1-\bx_2||\ll$, in a semantic sense) such that their label $y_1 \neq y_2$.
All possible counterfactuals to a given example $\bx_1$ constitute a distribution. We assume the availability of samples from it, forming pairs such as $\{(\bx_1,y_1),(\bx_2,y_2)\}$.
The counterfactual relation is undirected.

\vspace{3pt}\paragraph{Obtaining pairs of counterfactual examples.}
Some existing datasets explicitly contain pairs of counterfactual examples~\cite{baradel2019cophy,camburu2018snli,kaushik2019learning,park2019robust,park2019viewpoint}.
For example, \cite{kaushik2019learning} contains sentences (movie reviews) with positive and negative labels.
Annotators were instructed to edit a set of sentences to flip the label, thus creating counterfactual pairs (see examples in \fig\ref{figTeaser}).
Existing works simply use these as additional training point.
Our contribution is to use the \emph{relation} between these pairs, which is usually discarded.
In other datasets, counterfactual examples can be created by masking parts of the input, thus creative negative examples.
In \sect\ref{sectApplications}, we apply this approach to the COCO and VQA v2 datasets.

\subsection{Gradient supervision}
\label{tech}

\begin{figure}[t]
	\centering
	\includegraphics[width=.70\linewidth]{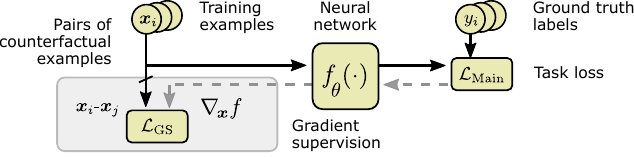}
	\caption{The proposed gradient supervision (GS) is an auxiliary loss on the gradient of a neural network with respect to its inputs, which is simply computed by backpropagation (dashed lines).
	Supervision for this gradient is generated from pairs of training examples identified as counterfactuals of one another. The loss $\mathcal{L}_\textrm{GS}$ is a cosine distance that encourages the gradient of the network to align with the vector between pairs of counterfactual examples.\label{figMethod}}
	\vspace{-8pt}
\end{figure}

To exploit relations between counterfactual examples, we introduce an auxiliary loss that supervises the gradient of the network $f_\btheta$.
We denote the gradient of the network with respect to its input at a point $\bx_i$ with 
${{\bg}}_i = \gradf(\bx_i)$.
Our new gradient supervision (GS) loss encourages ${{\bg}}_i$ 
to align with a ``ground truth'' gradient vector $\hat{\bg}_i$:
\begin{equation}
	\mathcal{L}_\mathrm{GS}\big({{\bg}}_i, \hat{\bg}_i \big) ~=~
	1 \;-\; \big( {{\bg}}_i . \hat{\bg}_i \big) \, / \,
	\big( || {{\bg}}_i || \, ||\hat{\bg}_i|| \big) ~.
	\label{eqLossGs}
\end{equation}
This definition is a cosine distance between ${{\bg}}_i$ and $\hat{\bg}_i$.
Assuming $\{(\bx_i,{y}_i),(\bx_j,{y}_j)\}$ is a pair of counterfactual examples, a ``ground truth'' gradient at $\bx_i$ is obtained as $\hat{\bg}_i=\bx_j$-$\bx_i$.
This represents the translation in the input space that should change the network output from ${y}_i$ to ${y}_j$.
Minimizing \eq\ref{eqLossGs} encourages the network's gradient to align with this vector at the training points.
Assuming $f$ is continuously differentiable, it also constrains the shape of $f$ \emph{between} training points.
This makes $f$ more faithful to the generating process behind the training data (see \sect\ref{sectTheory}).
Also note that the GS loss uses a local linearization of the network.
Although deep networks are highly non-linear globally, first-order approximations have found multiple uses, for example in providing explanations~\cite{ribeiro2016should,selvaraju2016grad} and generating adversarial examples~\cite{goodfellow2014explaining}.
In our application, this approximation is reasonable since pairs of counterfactual examples lie close to one another and to the classification boundary, by definition.

The network is optimized for a combination of the main and GS losses, $\mathcal{L} = \mathcal{L}_\mathrm{Main} + \lambda \, \mathcal{L}_\mathrm{GS}$, where $\lambda$ is a scalar hyperparameter.
The optimization of the GS loss requires backpropagating second-order derivatives through the network.
The computational cost over standard supervised training is of two extra backpropagations through the whole model for each mini-batch.

\vspace{3pt}\paragraph{Multiclass output.} In cases where the network output $\by$ is a vector, a ground truth gradient is only available for classes for which we have positive examples.
Denoting such a class \textit{gt}, we apply the GS loss only on the gradient of this class, using
${{\bg}}_i = \nabla_\bx f_i (\bx_i)$.
If a softmax is used, the output for one class depends on that of the others, so the derivative of the network is taken on its logits to make it dependent on one class only.

\subsection{How gradient supervision improves generalization}
\label{sectTheory}

\begin{figure}[t!]
	\centering
	\scriptsize Standard task loss\hspace{15mm}With gradient supervision\hspace{9mm}~\vspace{3pt}\\
	\includegraphics[width=.7\linewidth]{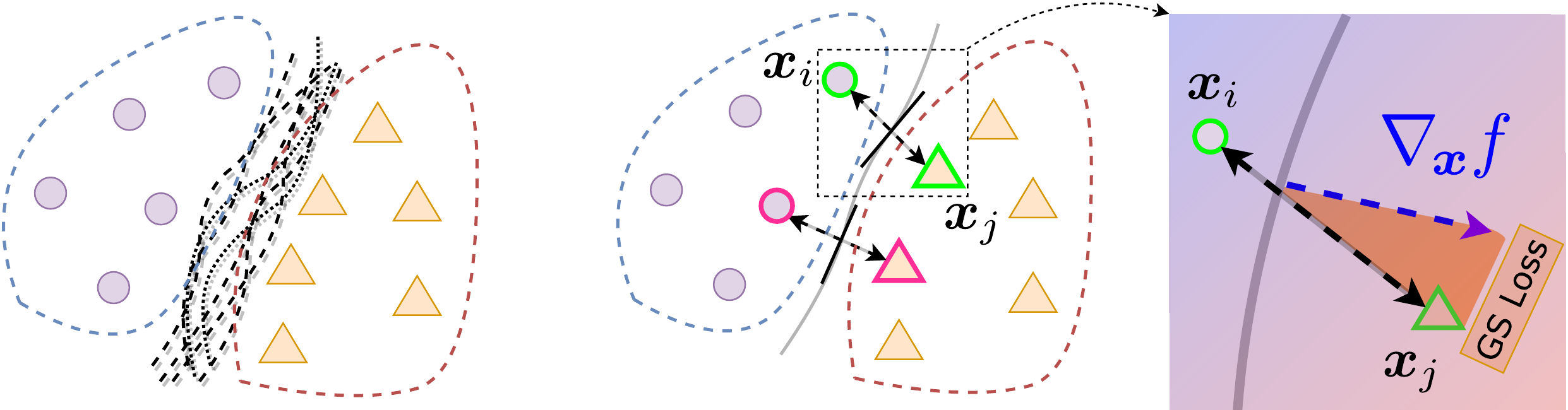}
	\vspace{-8pt}
	\caption{The proposed gradient supervision constrains the geometry of a model's decision boundary between classes.
	(Left)~We show possible decision boundaries consistent with a conventional supervised loss for two classes (circles and triangles representing training points of each).
	(Right)~The gradient supervision uses pairs of counterfactual examples $(\bx_i,\bx_j)$ to constrain classifier $f$ such that its local gradient aligns with the vector between these points.
	When the gradient supervision counterfactuals are included with the GS loss, the boundary is clearer between two classes. On the right, we show the GS loss for a pair of counterfactuals.
	\label{figPrinciple}}
	\vspace{-11pt}
\end{figure}

By training a machine learning model $f_\theta$, we seek to approximate an ideal $\mathcal{F}$ that represents the real-world process attributing the correct label $y=\mathcal{F}(\bx)$ to any possible input $\bx$.
Let us considering the Taylor expansion of  $f$ at a training point $\bx_j$:
\begin{equation}
f(\bx_j) ~=~
f(\bx_i) \;+\;
f'(\bx_i) \, (\bx_i-\bx_j) \;+\;
\underbrace{
	\frac{1}{2} \, f''(\bx_i) \, (\bx_i-\bx_j)^2 \;+\; \ldots
}_{\approx 0}
\label{eq:taylor}
\end{equation}
Our definition of a pair of counterfactual examples ($\bx_i$, $\bx_j$) (\sect\ref{sectMotivation}) implies that  $(\bx_i$-$\bx_j)^m$ approaches 0 ($m>1$).
For such a pair of nearby points, the terms beyond the first order virtually vanish.
It follows that the distance between $f(\bx_j)$ and $f(\bx_i)$ is maximized when the dot product $\gradf(\bx_i)\,.\,(\bx_i$-$\bx_j)$ is maximum.
This is precisely the desired behavior of $f$ in the vicinity of $\bx_i$ and $\bx_j$, since their ground truth labels $y_i$ and $y_j$ are different by our definition of counterfactuals.
This leads to the definition of the GS loss in \eq\ref{eqLossGs}.
Geometrically, it encourages the gradient of $f$ to align with the vector pointing from a point to its counterfactual, as illustrated in \fig\ref{figPrinciple}.

The conventional empirical risk minimization with non-convex functions leads to large numbers of local minimas. They correspond to multiple plausible decision boundaries with varied capability for generalization.
Our approach essentially modifies the optimization landscape for the parameters $\theta$ of $f$ such that the minimizer found after training is more likely to reflect the ideal~$\mathcal{F}$.

\section{Applications}
\label{sectApplications}

The proposed method is applicable to datasets with counterfactual examples in the training data.
They are sometimes provided explicitly~\cite{baradel2019cophy,camburu2018snli,kaushik2019learning,park2019robust,park2019viewpoint}.
Most interestingly, we show that they can be also be generated from existing annotations~\cite{das2016attention,goyal2017making,lin2014microsoft,suhr2017corpus,suhr2018corpus,vo2019composing}.

We selected four classical tasks in vision and language that are notoriously subject to poor generalization due to dataset biases.
Our experiments aim (1)~to measure the impact of gradient supervision on performance for well-known tasks, and (2)~to demonstrate that the necessary annotations are available in a variety of existing datasets.
We therefore prioritized the breadth of experiments and the use of simple models (details in supp. mat.) rather than chasing the state of the art on any particular task.
The method should readily apply to more complex models for any of these tasks.


\subsection{Visual question answering}

The task of visual question answering (VQA) involves an image and a related question, to which the model must determine the correct answer among a set of approximately 2,000 candidate answers.
Models trained on existing datasets (\eg VQA v2~\cite{goyal2016balanced}) are notoriously poor at generalization because of dataset biases.
These models rely on spurious correlations between the correct answer and certain words in the question.
We use the training/test splits of VQA-CP~\cite{agrawal2018don} that were manually organized such that the correlation between the questions' prefixes (first few words) and answers differ at training/test time.
Most methods evaluated on VQA-CP \textbf{use the explicit knowledge of this fact}~\cite{agrawal2018don,cadene2019rubi,clark2019don,grand2019adversarial,ramakrishnan2018overcoming,teney2019actively,teney2020unshuffling,yang2015stacked} or even of the ground truth set of prefixes, which defeats the purpose of evaluating generalization.
As discussed in the introduction, strong background assumptions are one of the two options to improve generalization beyond a set of labels.
Our method, however, follows the other option of using a different type of data, and \textbf{does not rest on the knowledge of the construction of VQA-CP}.

\vspace{3pt}\paragraph{Generating counterfactual examples.}
We build counterfactual examples for VQA-CP using annotations of human attention from~\cite{das2016attention}.
Given a question/image/answer triple $(\bq,\bI,\ba)$, we build its counterfactual counterpart $(\bq,\bI',\ba')$ by editing the image and answer.
The image $\bI$ is a set of features pre-extracted with a bottom-up attention model~\cite{anderson2017features} (typically a matrix of dimensions $N\times2048$).
We build $\bI'$ ($N'\times2048$, $N'\leq N$) by masking the features whose bounding boxes overlap with the human attention map past a certain threshold (details in supp. mat.).
The vector $\ba$ is a binary vector of correct answers over all candidates. We simply set all entries in $\ba'$ to zero.

\begin{table}[t!]
\scriptsize
\renewcommand{\tabcolsep}{0.01em}
\renewcommand{\arraystretch}{1.1}
\centering
\renewcommand{\tablefont}{\scriptsize}
	\begin{tabularx}{\linewidth}{X  cccc c cccc c cccc}
		\Xhline{1\arrayrulewidth}
							& \multicolumn{4}{c}{Val.} & ~ & \multicolumn{4}{c}{Test} & ~ & \multicolumn{4}{c}{Test ``focused''} \\
							\cline{2-5} \cline{7-10} \cline{12-15}
		Test data	$\rightarrow$	& All & YesNo & Nb & Other & ~ & All & YesNo & Nb & Other & ~ & All & YesNo & Nb & Other \\
		\Xhline{1\arrayrulewidth}
		SAN~\cite{yang2015stacked} & -- & -- & -- & -- & ~ & 25.0 & 38.4 & 11.1 & 21.7 & ~ & -- & -- & -- & -- \\
		GVQA~\cite{agrawal2018don} & -- & -- & -- & -- & ~ & 31.3 & 58.0 & 13.7 & 22.1 & ~ & -- & -- & -- & -- \\
		UpDown~\cite{teney2017challenge} & -- & -- & -- & -- & ~ & 39.1 & 62.4 & 15.1 & 34.5\\
		Ramakrishnan \etal, 2018~\cite{ramakrishnan2018overcoming} & -- & -- & -- & -- & ~ & 42.0 & \textbf{65.5} & 15.9 & 36.6 & ~ & -- & -- & -- & -- \\
		Grand and Belinkov, 2019~\cite{grand2019adversarial} & -- & -- & -- & -- & ~ & 42.3 & 59.7 & 14.8 & 40.8 & ~ & -- & -- & -- & -- \\
		RUBi~\cite{cadene2019rubi} & -- & -- & -- & -- & ~ & 47.1 & 68.7 & 20.3 & 43.2 & ~ & -- & -- & -- & -- \\
		Teney \etal, 2019 \cite{teney2019actively} & -- & -- & -- & -- & ~ & 46.0 & 58.2 & \textbf{29.5} & 44.3 & ~ & -- & -- & -- & -- \\
		Unshuffling~\cite{teney2020unshuffling} & -- & -- & -- & -- & ~ & 42.39 & 47.72 & 14.43 & 47.24 & ~ & -- & -- & -- & -- \\
		\Xhline{1\arrayrulewidth}
		Strong baseline~\cite{teney2020unshuffling} + CF data & 63.3 & 79.4 & 45.5 & 53.7 &  ~ & 46.0 & 61.3 & 15.6 & \textbf{46.0} &  ~ & 44.2 & 57.3 & 9.2 & \textbf{42.2} \\ 
		\textbf{+ CF data + GS}	& 62.4 & 77.8 & 43.8 & 53.6 &  ~ & \textbf{46.8} & 64.5 & 15.3 & 45.9 &  ~ & \textbf{46.2} & \textbf{63.5} & \textbf{10.5} & 41.4 \\ 
		\Xhline{1\arrayrulewidth}
		\multicolumn{2}{l}{Weak baseline (BUTD \cite{teney2017challenge}),}\\
		trained on `Other' only & -- & -- & -- & 54.7 & ~ & -- & -- & -- & 43.3 & ~ & -- & -- & -- & 40.6 \\ 
		+ CF data & -- & -- & -- & 55.9 & ~ & -- & -- & -- & 45.0 & ~ & -- & -- & -- & 40.6 \\ 
		\textbf{+ CF data + GS} & -- & -- & -- & 56.1 & ~ & -- & -- & -- & 44.7 & ~ & -- & -- & -- & 38.3 \\ 
		\Xhline{1\arrayrulewidth}
	\end{tabularx}

	\normalsize
	\vspace{4pt}
	\normalsize
	\caption{Application to VQA-CP v2. Existing methods all rely on built-in knowledge of the construction procedure of the dataset, defeating some of the claimed improvements in robustness. Using counterfactual data with the proposed gradient supervision (GS) improve performance on most question types on the out-of-distribution test sets (see text for discussion).\label{tabVqa}}
	\vspace{-10pt}
\end{table}

\vspace{3pt}\paragraph{Experimental setting.}
For training, we use the training split of VQA-CP, minus 8,000 questions held out as an ``in-domain'' validation set~(as in \cite{teney2020unshuffling}).
We generate counterfactual versions of the training examples that have a human attention map (approx. 7\% of them).
For evaluation, we use (1)~our ``in-domain'' validation set (held out from the training set), (2)~the official VQA-CP test set (which has a different correlation between prefixes and answers), and (3)~a new \textit{focused} test set.

The \textit{focused} test set contains the questions from VQA-CP test from which we only keep image features of regions looked at by humans to answer the questions.
We essentially perform the opposite of the building of counterfactual examples, and mask regions where the human attention is \textbf{below} a low threshold.
Answering questions from the \textit{focused} test set should intuitively be easier, since the background and distracting image regions have been removed.
However, a model that relies on context (question or irrelevant image regions) rather than strictly on the relevant visual evidence will do poorly on the focused test set.
This serves to measure robustness beyond the question biases that VQA-CP was specifically designed for.

\vspace{3pt}\paragraph{Results}
We present results of our method applied on top of two existing models.
The first (\textit{weak baseline}) is the popular BUTD model~\cite{anderson2017features,teney2017challenge}.
The second (\textit{strong baseline}) is the ``unshuffling'' method of \cite{teney2020unshuffling}, which was specifically tuned to address the language biases evaluated with VQA-CP.
We compare the baseline model with the same model trained with the additional counterfactual data, and then with the additional GS loss. The performance improves on most question types with each of these additions.
The ``focused'' provides an out-of-distribution evaluation complementary to the VQA-CP test set (which only accounts for language biases).
It shows the improvements expected from our method to a larger extent that the VQA-CP test set.
This suggests that evaluating generalization in VQA is still not completely addressed with the current benchmarks.
Importantly, the improvements over both the weak and strong baselines indicate that \textbf{the proposed method is not redundant with existing methods that specifically address the language biases measured by VQA-CP}, like the strong baseline.
Additional details are provided in the supplementary material.

\subsection{Multi-label image classification}
We apply our method to the COCO dataset~\cite{lin2014microsoft}.
Its images feature objects from 80 classes.
They appear in common situations such that the patterns of co-occurrence are highly predictable: a bicycle often appears together with a person, and a traffic light often appears with cars, for example.
These images serve as the basis of a number of benchmarks for image detection~\cite{lin2014microsoft}, captioning~\cite{chen2015microsoft}, visual question answering~\cite{antol2015vqa}, etc.
They all inherit the biases inherent to the COCO images~\cite{agrawal2018don,hendricks2018women,wang2019balanced} which is an increasing cause of concern.
A method to improve generalization in this context has a wide potential impact.

\vspace{3pt}\paragraph{Experimental setting.}
We consider a simple multi-label classification task that captures the core issue of dataset biases that affect higher-level tasks (captioning for example~\cite{hendricks2018women}).
Each image is associated with a binary vector of size 80 that represents the presence of at least one object of the corresponding class in the image.
The task is to predict this binary vector. Performance is measured with the mean average precision (mAP) over all classes.
The model is a feed-forward neural network that performs an 80-class binary classification with sigmoid outputs, over pre-extracted ResNet-based visual features.
We pre-extract these features with the bottom-up attention model of Anderson \etal~\cite{anderson2017features}.
They are spatially pooled into a 2048-dimensional vector.
The model is trained with a standard binary cross-entropy loss (details in the supplementary material).

\begin{table}[t!]
\scriptsize
\renewcommand{\tabcolsep}{0.70em}
\renewcommand{\arraystretch}{1.01}
\centering
\renewcommand{\tablefont}{\footnotesize}
	\footnotesize
	\begin{tabularx}{\linewidth}{Xccc}
	\Xhline{1\arrayrulewidth}
						& \multicolumn{3}{c}{\tablefont COCO Multi-label classification} \\ \cline{2-4}
						& Original & Edited images & Hard edited \\
	Test data	$\rightarrow$	& images & images & images\\
	\Xhline{1\arrayrulewidth}
	Random predictions (chance) & 5.1 & 3.9 & 7.8 \\
	Baseline w/o edited tr. examples & 71.8 & 58.1 & 54.8 \\
	Baseline w/ edited tr. examples & 72.1 & 64.0 & 56.0 \\
	+ \textbf{GS, counterfactual relations} & \textbf{72.9} & \textbf{65.2} & \textbf{57.7} \\
	+ GS, random relations & 71.8 & 63.9 & 56.1 \\
	\Xhline{1\arrayrulewidth}
	\end{tabularx}\\
	\normalsize
	\vspace{4pt}
	\normalsize
	\caption{Application to multi-label classification on COCO.
	We use counterfactual examples generated by masking objects with the inpainter GAN~\cite{agarwal2019towards,shetty2018adversarial}.
	Our method allows to train a model that is less reliant less on common object co-occurrences of the training set.
	The most striking improvements are measurable with images that feature sets of objects that appear rarely (``Edited'') or never (``Hard edited'') during training.\label{tabCoco}}
	\vspace{-10pt}
\end{table}

\begin{figure}[ht!]
	\centering
	\hangBoxC{\includegraphics[width=.16\linewidth]{COCO_val2014_000000039659_000000000048.jpg}}~\hangBoxC{\includegraphics[width=.16\linewidth]{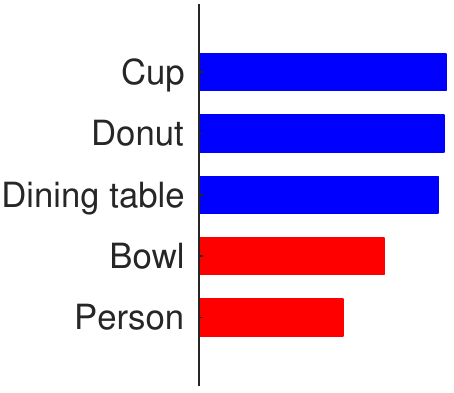}}\hangBoxC{\includegraphics[width=.16\linewidth]{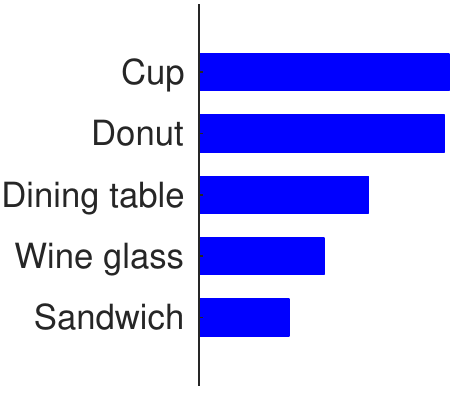}}~~
	\hangBoxC{\includegraphics[width=.16\linewidth]{COCO_val2014_000000120420.jpg}}~\hangBoxC{\includegraphics[width=.16\linewidth]{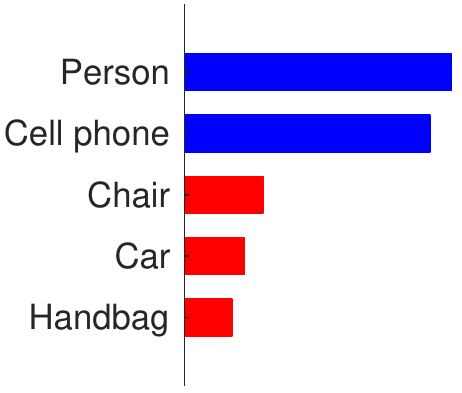}}\hangBoxC{\includegraphics[width=.16\linewidth]{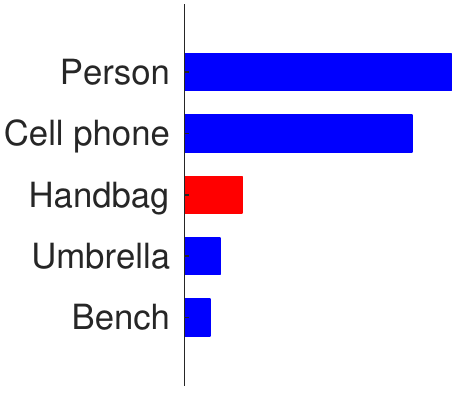}}\\
	\hangBoxC{\includegraphics[width=.16\linewidth]{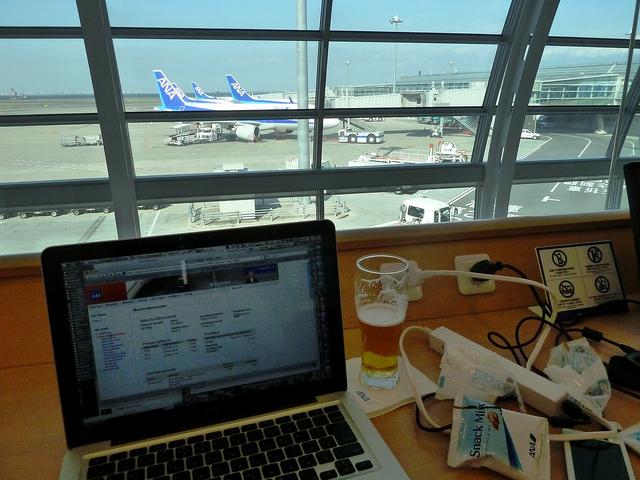}}~\hangBoxC{\includegraphics[width=.16\linewidth]{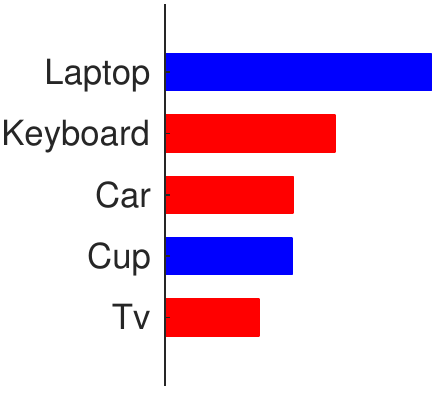}}\hangBoxC{\includegraphics[width=.16\linewidth]{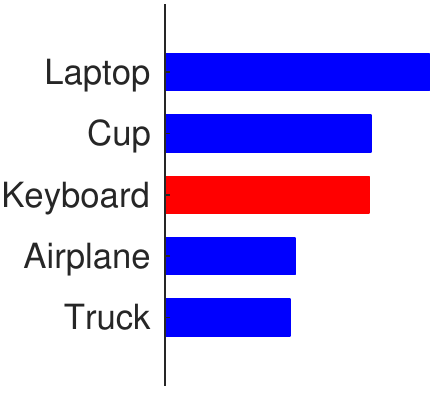}}~~
	\hangBoxC{\includegraphics[width=.16\linewidth]{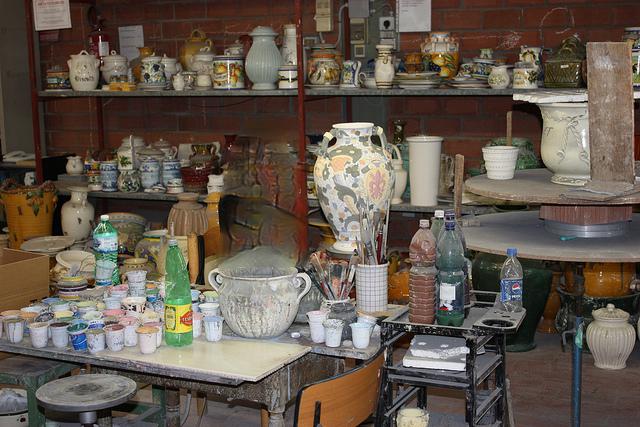}}~\hangBoxC{\includegraphics[width=.16\linewidth]{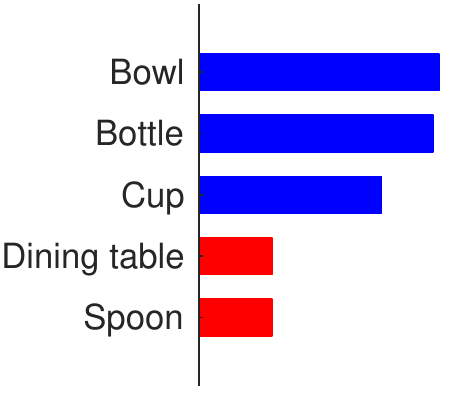}}\hangBoxC{\includegraphics[width=.16\linewidth]{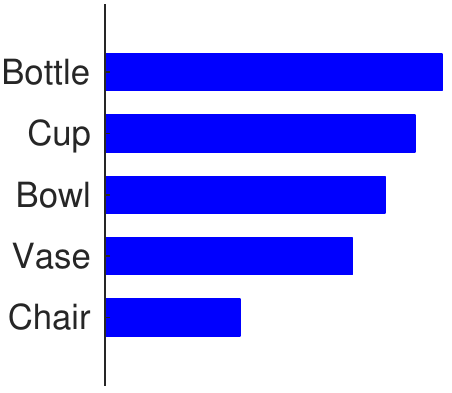}}\\
	\hangBoxC{\includegraphics[width=.16\linewidth]{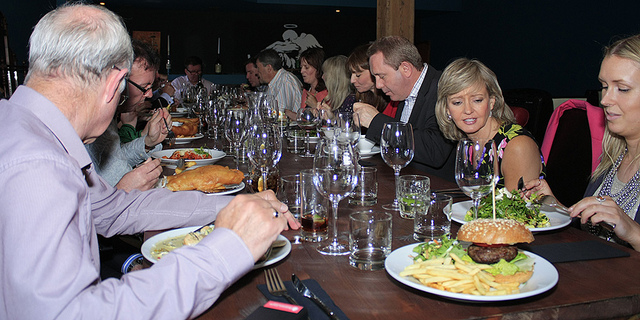}}~\hangBoxC{\includegraphics[width=.16\linewidth]{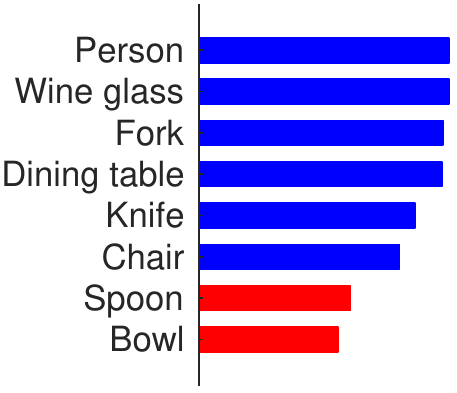}}\hangBoxC{\includegraphics[width=.16\linewidth]{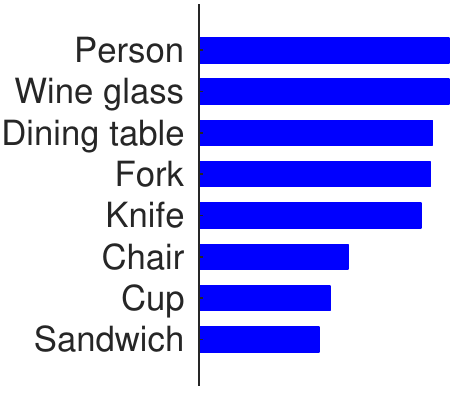}}~~
	\hangBoxC{\includegraphics[width=.16\linewidth]{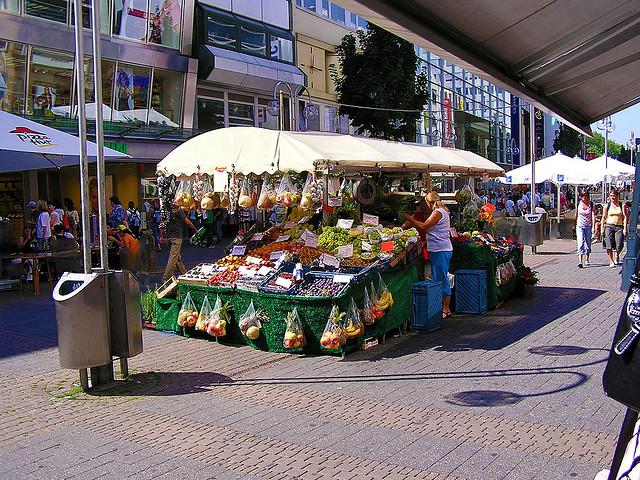}}~\hangBoxC{\includegraphics[width=.16\linewidth]{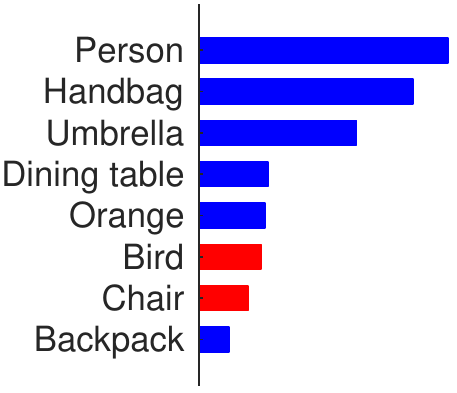}}\hangBoxC{\includegraphics[width=.16\linewidth]{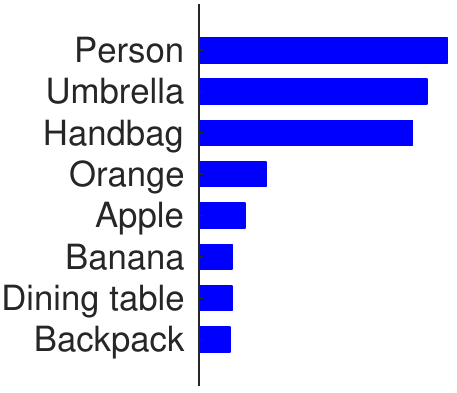}}\\
	\hangBoxC{\includegraphics[width=.16\linewidth]{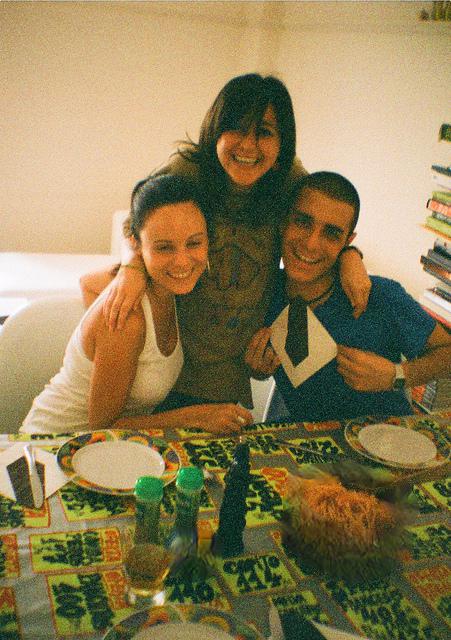}}~\hangBoxC{\includegraphics[width=.16\linewidth]{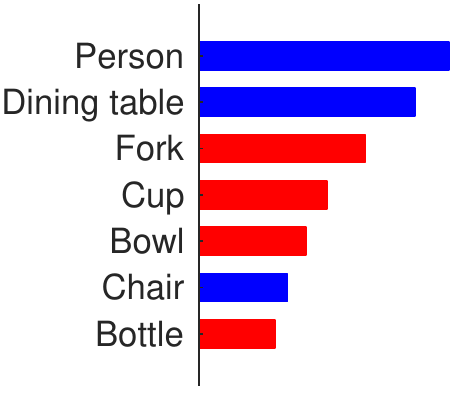}}\hangBoxC{\includegraphics[width=.16\linewidth]{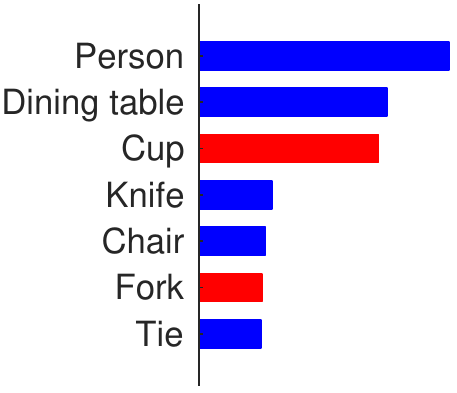}}~~
	\hangBoxC{\includegraphics[width=.16\linewidth]{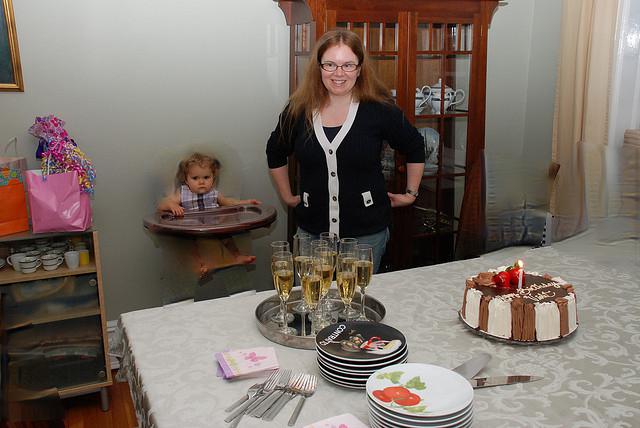}}~\hangBoxC{\includegraphics[width=.16\linewidth]{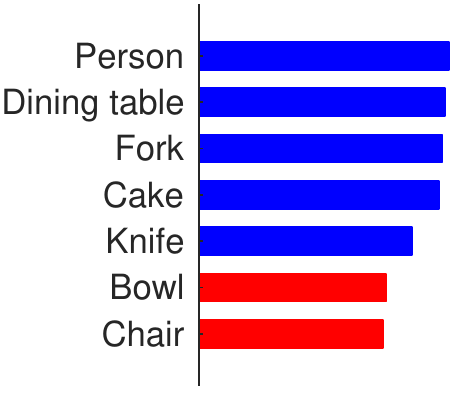}}\hangBoxC{\includegraphics[width=.16\linewidth]{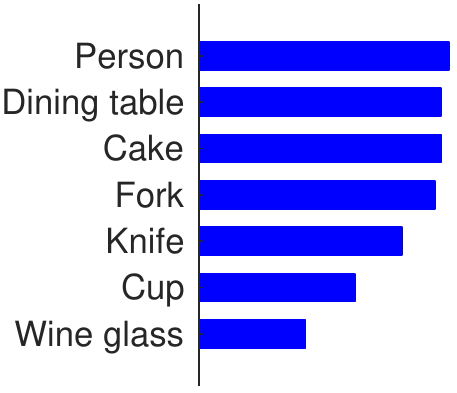}}\\
	\vspace{-7pt}
	\caption{Qualitative examples of multi-label classification on COCO.
	We show the input image and the scores of the top predicted labels by the baseline and by our method (blue: correct, red: incorrect).
	The baseline can erroneously predict common co-occurring objects, such as a \textit{person} with food items (top left) even though there is no visual evidence for the former.
	Our method is better at predicting unusual combinations, such as as a \textit{donut} with a \textit{wineglass}~(first row, left) or a \textit{laptop} with an \textit{airplane} (second row, left).\label{figCoco}}
	\vspace{-9pt}
\end{figure}

\vspace{3pt}\paragraph{Generating counterfactual examples.}
Counterfactual examples can be generated using existing annotation in COCO. 
Agarwal \etal~\cite{agarwal2019towards} used the inpainter GAN~\cite{shetty2018adversarial} to edit images by masking selected objects.
This only requires the original labels and bounding boxes.
The edited images represent a ``minimal change'' that makes the corresponding label negative, which agrees with our definition of counterfactuals.
The vector of ground truth labels for edited images are edited accordingly.
For training, we use all images produced by \cite{agarwal2019towards} from the COCO \textit{train2014} split (original and edited versions).
For evaluation, we use their images from the \textit{val2014} split (original and edited version, evaluated separately).
We also create an additional evaluation split named ``Hard edited images''. It contains a subset of edited images with patterns of classes that \textbf{never} appear in the training set.

\vspace{3pt}\paragraph{Results.}
We first compare the baseline model trained with the original images only, and then with the original and edited images (Table~\ref{tabCoco}).
The performance improves (71.8$\rightarrow$72.1\%), which is particularly clear when evaluated on edited images (58.1$\rightarrow$64.0\%).
This is because the patterns of co-occurrence in the training data cannot blindly relied on with the edited images.
The images in this set depict situations that are unusual in the training set, such as a surfboard without a person on top, or a man on a tennis court who is not holding a racquet.
A model that relies on common co-occurrences in the training set rather than strictly on visual evidence can do well on the original images, but not on edited ones.
An improvement from additional data is not surprising. It is still worth emphasizing that the edited images were generated ``for free'' using existing annotations in COCO.

Training the model with the proposed gradient supervision (GS) further improves the precision (72.1$\rightarrow$72.9\%).
This is again more significant on the edited images (64.0$\rightarrow$65.2\%).
The improvement is highest on the set of ``hard edited images'' (56.0$\rightarrow$57.7\%).
As an ablation, we train the GS model with random pairwise relations instead of relations between counterfactual pairs.
The performance is clearly worse, showing that the value of GS is in leveraging an additional training signal, rather than setting arbitrary constraints on the gradient like existing unsupervised regularizers~\cite{lirobust,jakubovitz2018improving,woods2019adversarial}.
In \fig\ref{figCoco}, we provide qualitative examples from the evaluation sets where the predictions of our model improve over the baseline.



\subsection{NLP Tasks: sentiment analysis and natural language inference}

\begin{table}[t!]
\scriptsize
\renewcommand{\tabcolsep}{0.15em}
\renewcommand{\arraystretch}{1.01}
\centering
\renewcommand{\tablefont}{\footnotesize}
	\footnotesize
	\begin{tabularx}{\linewidth}{Xcccccccc}
	\Xhline{1\arrayrulewidth}
						~ & \multicolumn{3}{c}{IMDb with counterfactuals} & ~ & \multicolumn{3}{c}{Zero-shot transfer} \\ 
	Test data	$\rightarrow$	~ & Val. & Test original & Test edited & ~ & Amazon & Twitter & Yelp \\
	\Xhline{1\arrayrulewidth}
	Random predictions (chance) & 51.4 & 47.7 & 49.2 & ~ & 47.3 & 53.3 & 45.4 \\
	Baseline w/o edited tr. data & 71.2 & 82.6 & 55.3 & ~ & 78.6 & 61.0 & 82.8\\
	Baseline w/ edited tr. data & 85.7& 82.0& 88.7 & ~ & 80.8 & 63.1 & 87.4 \\
	+ \textbf{GS}, counterfactual rel. & \textbf{89.8} & \textbf{83.8} & \textbf{91.2} & ~ & \textbf{81.6} & \textbf{65.4} & \textbf{88.8} \\
	+ GS, random relations & 50.8 & 49.2 & 52.0 & ~ & 47.4 & 61.2 & 57.4 \\ 
	\Xhline{1\arrayrulewidth}
	\end{tabularx}
	\vspace{6pt}

	\renewcommand{\tabcolsep}{0.20em}
	\begin{tabularx}{\linewidth}{Xcccccccc}
	\Xhline{1\arrayrulewidth}
						~ & \multicolumn{3}{c}{SNLI with counterfactuals} & ~ & \multicolumn{1}{c}{Zero-shot transfer} \\ 
	Test data	$\rightarrow$ 	~ & Val. & Test original & Test edited & ~ & MultiNLI dev. \\
	\Xhline{1\arrayrulewidth}
	Random predictions (chance) & 30.8 & 34.6 & 32.9 & ~ & 31.9 \\
	Baseline w/o edited tr. data & 61.8 & 42.0 & 59.0 & ~ & 46.0 \\
	Baseline w/ edited tr. data & 61.3 & 39.1 & 57.8 & ~ & 42.4 \\
	+ \textbf{GS}, counterfactual relations & \textbf{64.8} & \textbf{44.4} & \textbf{61.2} & ~ & \textbf{46.8} \\
	+ GS, random relations & 58.5 & 40.4 & 58.6 & ~ & 45.7 \\
	\Xhline{1\arrayrulewidth}
	\end{tabularx}

	\normalsize
	\vspace{4pt}
	\normalsize
	\caption{Application to sentiment analysis (top) and natural language inference (bottom), trained resp. on the subsets of the IMDb and SNLI datasets augmented with ``edited'' counterfactual examples~\cite{kaushik2019learning}.
	Our technique brings clear improvements over mere data augmentation baseline (accuracy in \%), in particular when evaluated on the edited test data (on which biases from the original training data cannot be relied on) and on test data from other datasets (no fine-tuning is used).\label{tabSentiment}\label{tabSnli}}
	\vspace{-10pt}
\end{table}

The task of \textbf{sentiment analysis} is to assign a positive or negative label to a text snippet, such as a movie or restaurant review.
For training, we use the extension of the IMDb dataset~\cite{maas2011learning} of movie reviews by Kaushik \etal~\cite{kaushik2019learning}.
They collected counterfactual examples by instructing crowdworkers to edit sentences from the original dataset to flip their label.
They showed that a standard model trained on the original data performs poorly when evaluated on edited data, indicating that it relies heavily on dataset biases (\eg the movie genre being predictive of the label).
They then used edited data during training (simply mixing it with the original data) and showed much better performance in all evaluation settings, even when controlling for the amount of additional training examples.
Our contribution is to use GS to leverage the relations between the pairs of original/edited examples.

The task of \textbf{natural language inference (NLI)} is to classify a pair of sentences, named the premise and the hypothesis, into $\{$\textit{entailment}, \textit{contradiction}, \textit{neutral}$\}$ according to their logical relationship.
We use the extension of the SNLI dataset~\cite{bowman2015large} by Kaushik \etal~\cite{kaushik2019learning}.
They instructed crowdworkers to edit original examples to change their labels.
Each original example is supplemented with versions produced by editing the premise or the hypothesis, to either of the other two classes.
The original and edited data together are therefore four times as large as the original data alone.

\vspace{3pt}\paragraph{Results on sentiment analysis.}
We first compare a model trained with the original data, and with the original and edited data as simple augmentation (Table~\ref{tabSentiment}).
The improvement is significant when tested on edited data~(55.3$\rightarrow$88.7\%).
We then train the model with our GS loss. The added improvement is visible on both the original data (82.0$\rightarrow$83.8\%) and on the edited data (88.7$\rightarrow$91.2\%).
The evaluation on edited examples is the more challenging setting, because spurious correlations from the original training data cannot be relied on.
The ablation that uses GS with random relations completely fails, confirming the value of the supervision with relations between pairs of related examples.

We additionally evaluate the model on out-of-sample data with three additional test sets: Amazon Reviews~\cite{ni-etal-2019-justifying}, Semeval 2017 (Twitter data)~\cite{rosenthal-etal-2017-semeval}, and Yelp reviews~\cite{yelpChallenge}.
The model trained on IMDb is applied \textbf{without any fine-tuning} to these, which constitutes a significant challenge in terms of generalization. We observe a clear gain over the data augmentation baseline on all three.

\vspace{3pt}\paragraph{Results on NLI.}
We perform the same set of experiments on NLI.
The fairest point of comparison is again the model trained with the original and edited data.
Using the GS loss on top of it brings again a clear improvement (Table~\ref{tabSnli}), both when evaluated on standard test data and on edited examples.
As an additional measure of generalization, we also evaluate the same models on the dev. set of MultiNLI~\cite{williams2017broad} without any fine-tuning.
There is a significant domain shift between the datasets. Using the edited examples for data augmentation actually hurts the performance here, most likely because they constitute very ``unnatural'' sentences, such that easy-to-pick-up language cues cannot be relied on.
Using GS (with always uses the edited data as augmentations as well) brings back the performance higher, and above the baseline trained only on the original data.

\begin{figure}[t!]
	\centering
	\includegraphics[width=.28\linewidth]{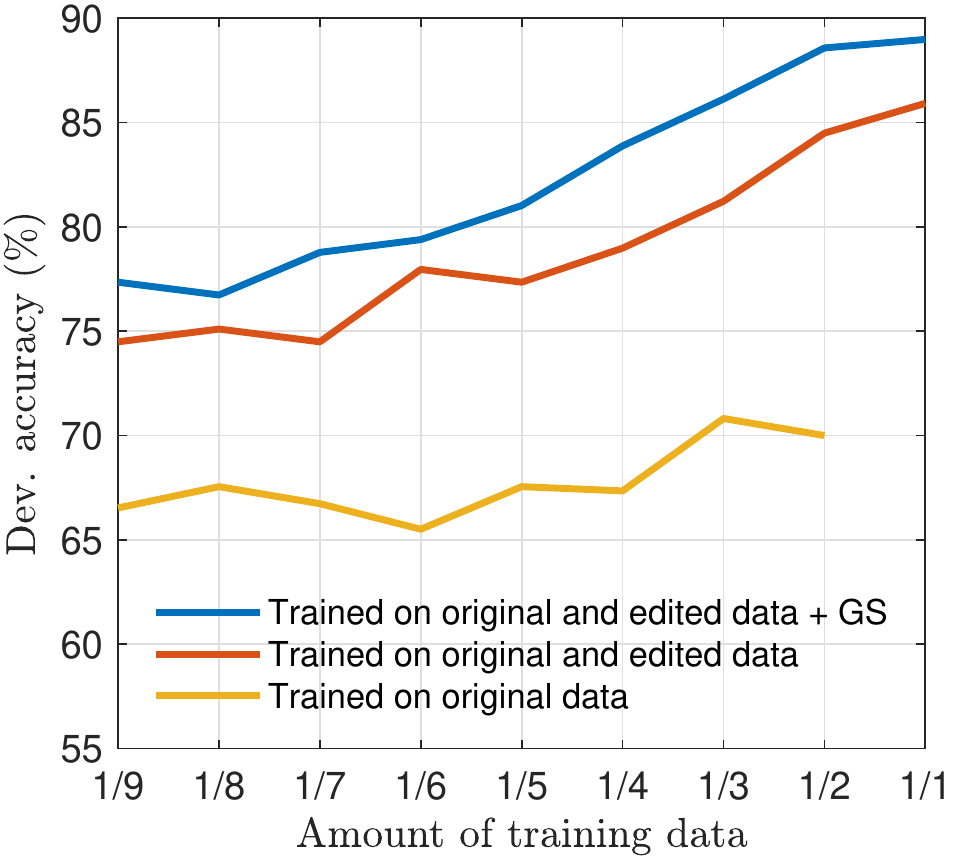}
	\label{figSentimentReducedData}
	\vspace{-5pt}
	\caption{Training on different amounts of data for sentiment analysis. The proposed gradient supervision (GS) brings a clear improvement over a model trained with edited examples for simple data augmentation, and even more so over a model trained with the same number of original training examples.}
	\vspace{-9pt}
\end{figure}

\vspace{3pt}\paragraph{Limitations.}
Our NLP experiments were conducted with simple models and relatively little data.
The current state of the art in sentiment analysis and NLI is achieved by transformer-based models~\cite{devlin2018bert} trained on vastly more data.
Kaushik \etal~\cite{kaushik2019learning} showed that counterfactual examples are much more valuable than the same amount of standard data, including for fine-tuning a BERT model for NLI.
The application of our technique to the extremely-large data regime, including with large-scale language models, is an exciting direction for future work.

\section{Conclusions}

We proposed a new training objective that improves the generalization capabilities of neural networks by supervising their gradient, and using an unused training signal found in many datasets.
While most machine learning models rely on identifying correlations between inputs and output, we showed that relations between counterfactual examples provide a fundamentally different, complementary type of information.
We showed theoretically and empirically that our technique can shape the decision boundary of the model to be more faithful to the causal mechanisms that generated the data.
Practically speaking, the model is then more likely to be ``right for the right reasons''.
We showed that this effect brings significant improvements on a number of tasks when evaluated with out-of-distribution test data.
We demonstrated that the required annotations can be extracted from existing datasets for a number of tasks.

There is a number of additional tasks and datasets on which our method can readily apply~\cite{baradel2019cophy,camburu2018snli,park2019robust,park2019viewpoint,suhr2017corpus,suhr2018corpus,vo2019composing}.
Scaling up the technique to state-of-the-art models in vision and NLP~\cite{devlin2018bert,he2015resnet,liu2019cbnet,su2019vl,tan2019lxmert} is another exciting direction for future work.


\clearpage
\bibliographystyle{splncs04}

\clearpage
\appendix

\section*{Supplementary material}
\vspace{12pt}


\section{Application to VQA}

\begin{figure}[p]
	\centering
	\renewcommand{\tabcolsep}{0.3em}
	\renewcommand{\arraystretch}{0.8}
	\begin{tabularx}{\linewidth}{cl}
	\hangBoxC{\includegraphics[height=.15\linewidth]{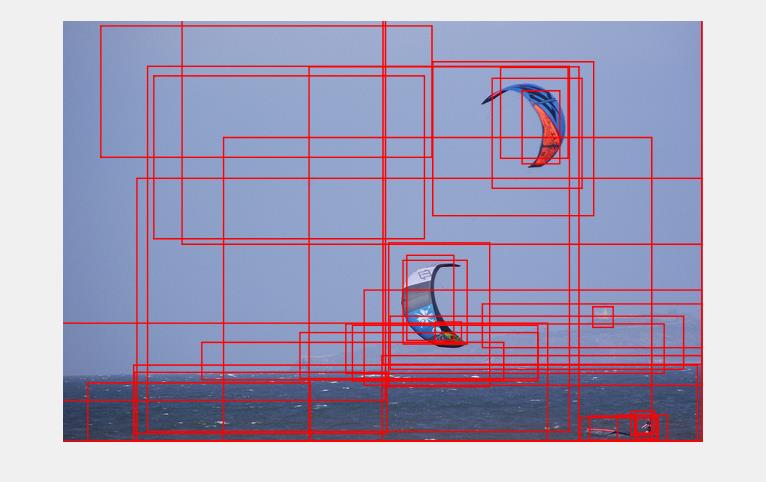}\includegraphics[height=.15\linewidth]{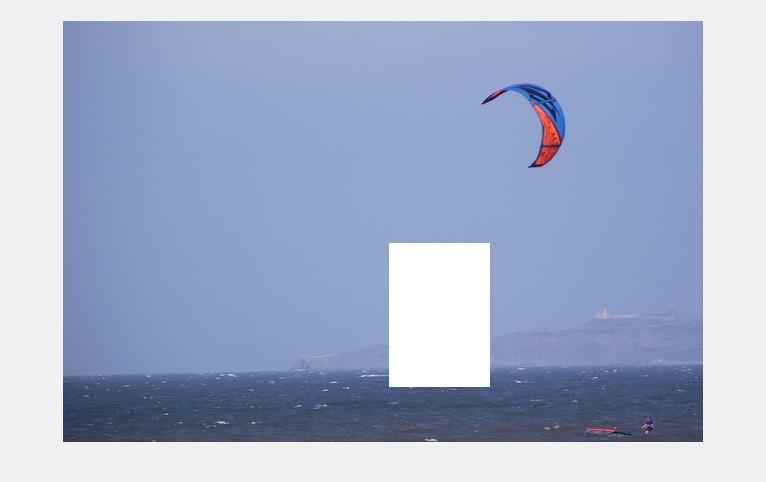}}&\hangBoxC{\footnotesize What is floating in the sky ?\\GT Answer(s): kites, kite, sail}\\
	\hangBoxC{\includegraphics[height=.15\linewidth]{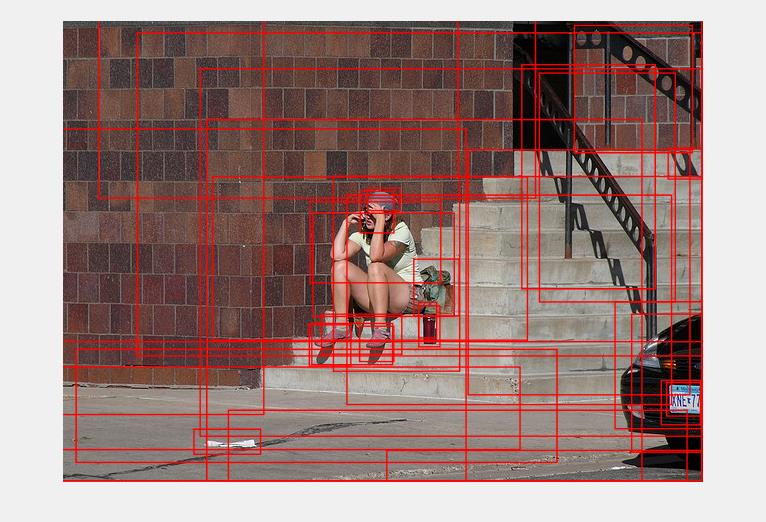}\includegraphics[height=.15\linewidth]{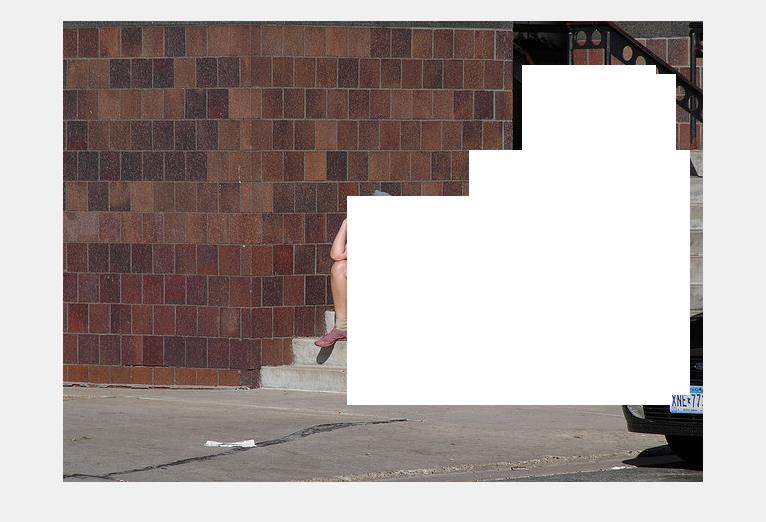}}&\hangBoxC{\footnotesize Where is the woman sitting ?\\GT Answer(s): stairs, steps}\\
	\hangBoxC{\includegraphics[height=.15\linewidth]{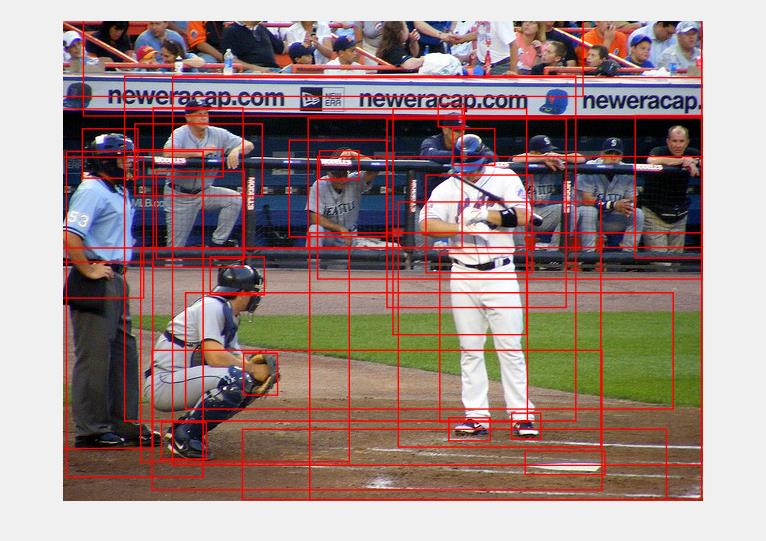}\includegraphics[height=.15\linewidth]{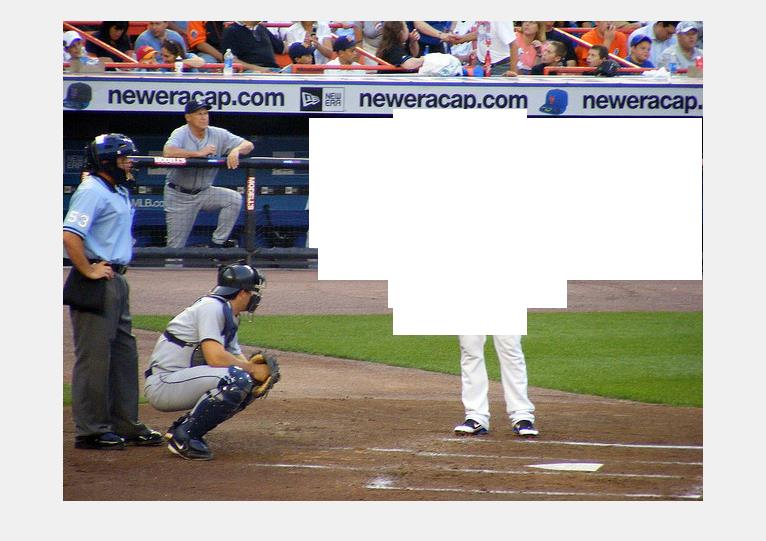}}&\hangBoxC{\footnotesize What team is the batter on ?\\GT Answer(s): white, yankees, mets, giants}\\
	\hangBoxC{\includegraphics[height=.15\linewidth]{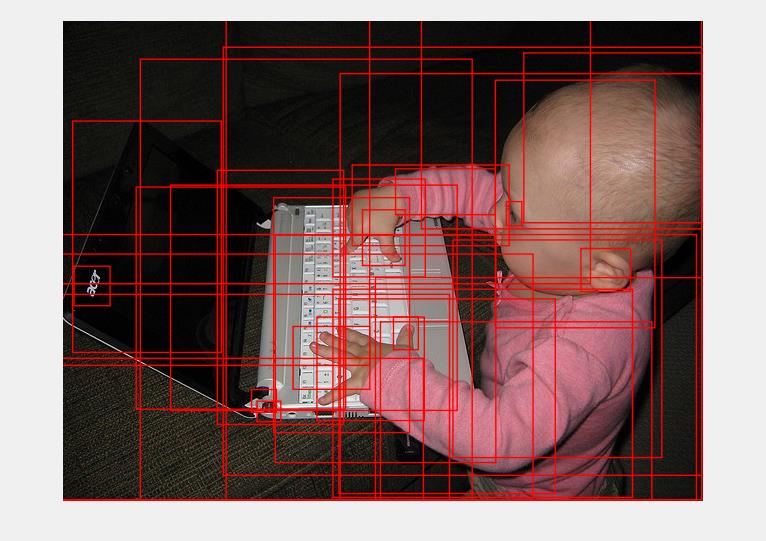}\includegraphics[height=.15\linewidth]{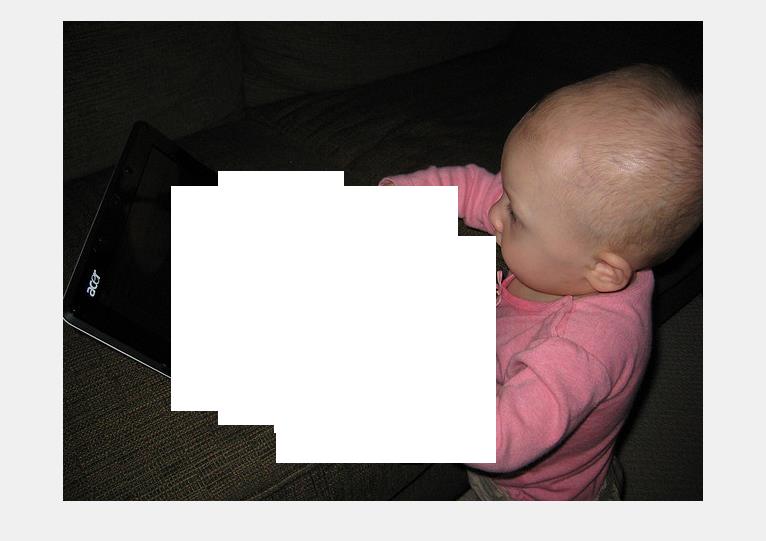}}&\hangBoxC{\footnotesize Where is the baby looking ?\\GT Answer(s): laptop, screen, monitor}\\
	\hangBoxC{\includegraphics[height=.15\linewidth]{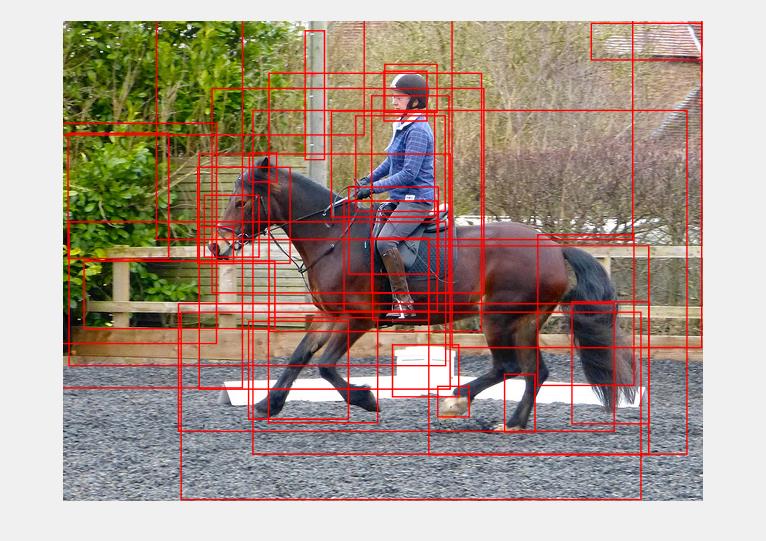}\includegraphics[height=.15\linewidth]{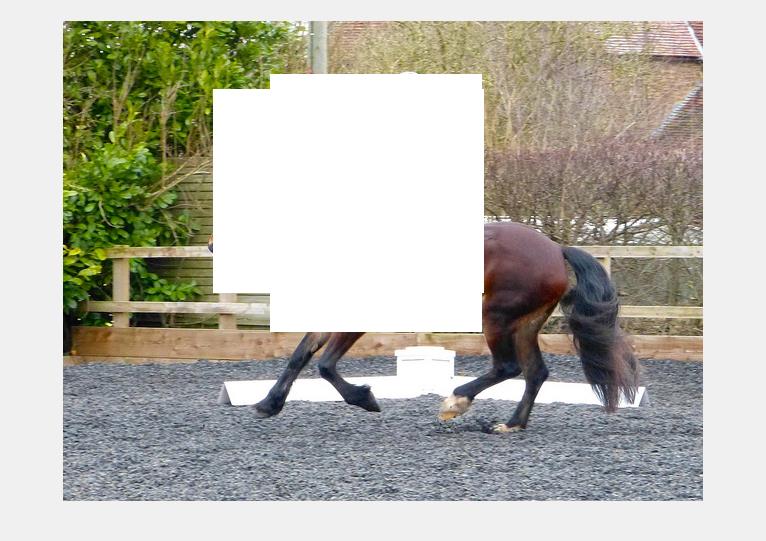}}&\hangBoxC{\footnotesize What is the sex of rider ?\\GT Answer(s): female, male}\\
	\hangBoxC{\includegraphics[height=.15\linewidth]{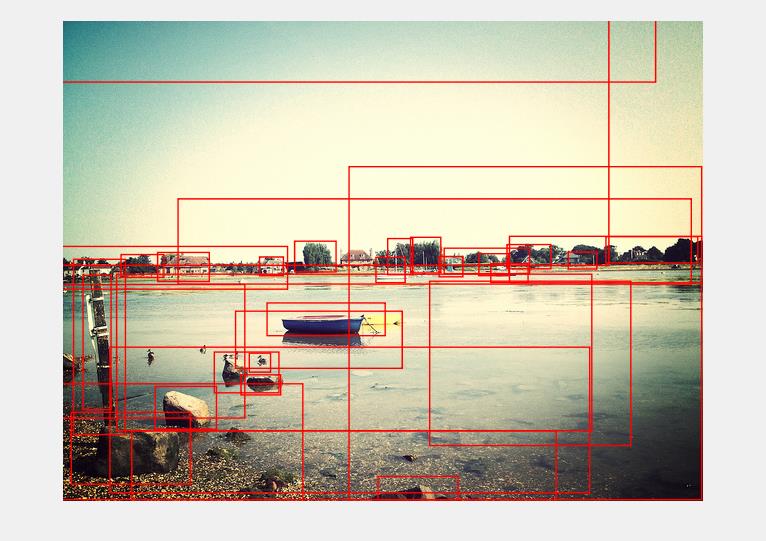}\includegraphics[height=.15\linewidth]{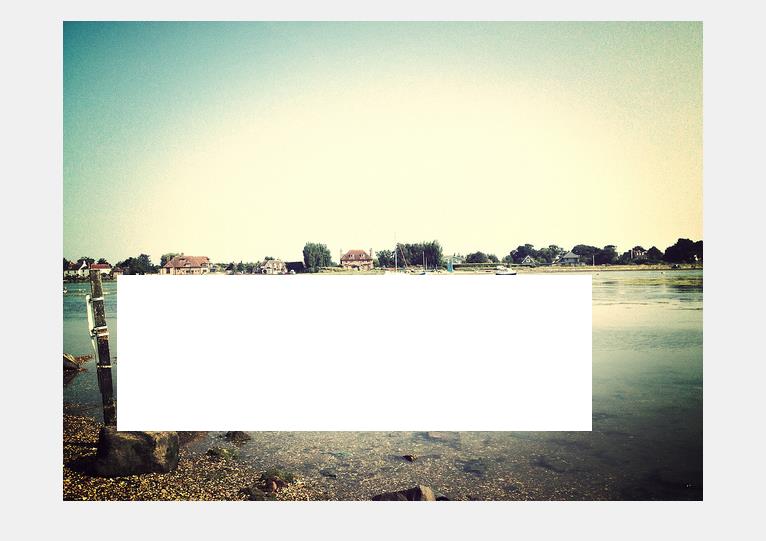}}&\hangBoxC{\footnotesize What kind of boat is on the water ?\\GT Answer(s): canoe, paddle}\\
	\hangBoxC{\includegraphics[height=.15\linewidth]{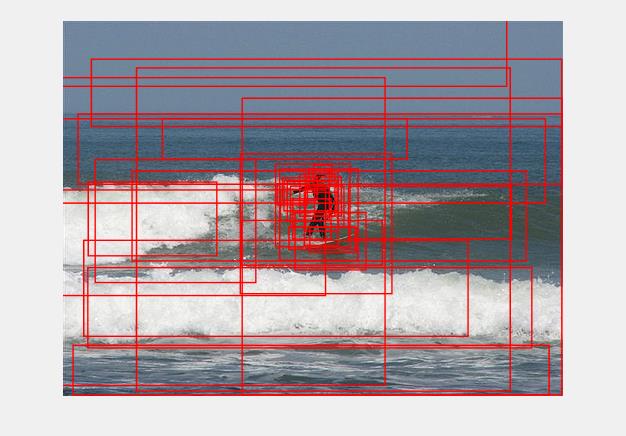}\includegraphics[height=.15\linewidth]{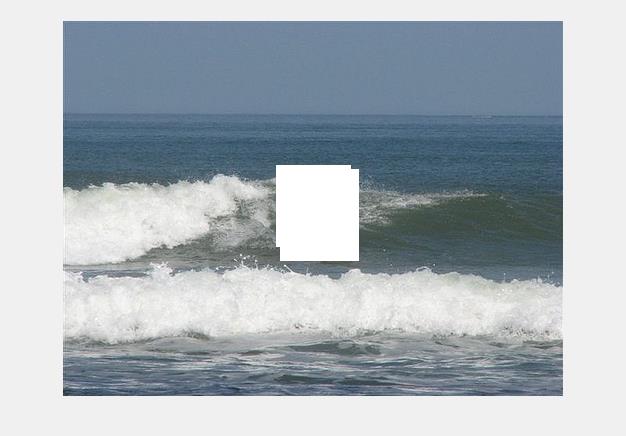}}&\hangBoxC{\footnotesize What sport is the person participating in ?\\GT Answer(s): surfing}\\
	\hangBoxC{\includegraphics[height=.15\linewidth]{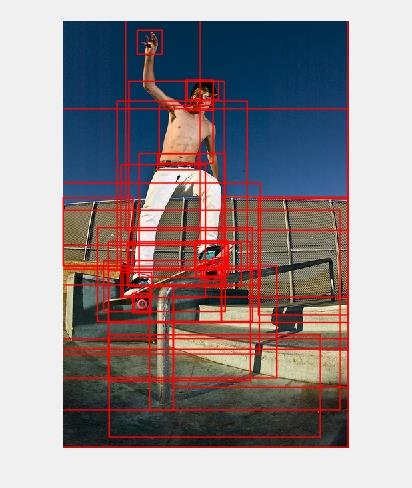}\includegraphics[height=.15\linewidth]{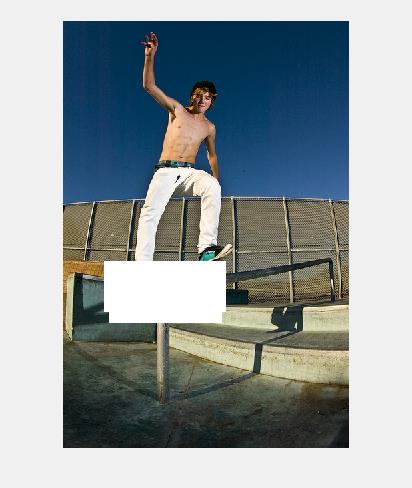}}&\hangBoxC{\footnotesize What is this person standing on ?\\GT Answer(s): skateboard}\\
	\hangBoxC{\includegraphics[height=.15\linewidth]{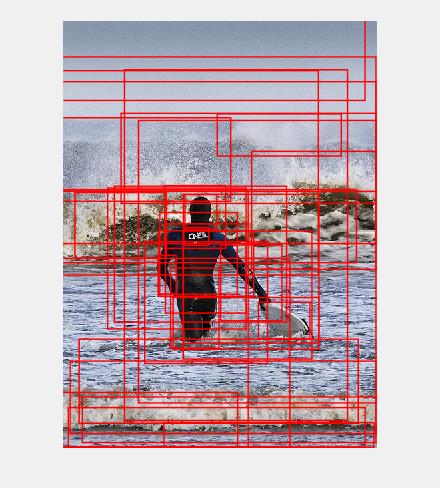}\includegraphics[height=.15\linewidth]{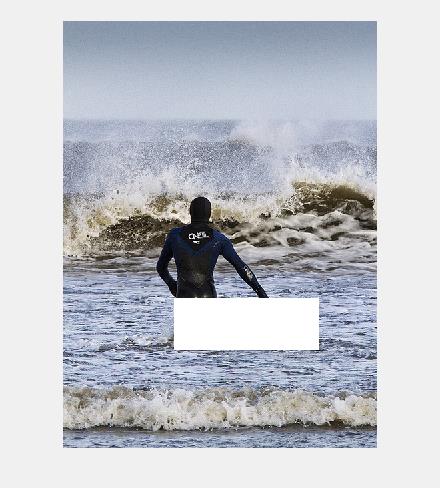}}&\hangBoxC{\footnotesize What is the person in photo holding ?\\GT Answer(s): surfboard}\\
	\end{tabularx}
	\vspace{-7pt}
	\caption{Application to VQA. Examples of original examples (with their ground truth answer) and their counterfactual version. Red boxes indicate regions that were candidates for masking when generating the counterfactual versions.\label{figSuppVqa}}
	\vspace{-9pt}
\end{figure}

\vspace{3pt}\paragraph{Data.}
We generate the counterfactual examples by masking image features on-the-fly, during training, according the the human attention maps of~\cite{das2016attention}. We use image features from \cite{anderson2017features}, which correspond to bounding boxes in the image. We mask the features whose boxes overlap with a fraction of the the human attention map above a fixed threshold. We use the precomputed overlap score from~\cite{selvaraju2019taking}, which is a scalar in $[0,1]$, and set the threshold at 0.2 (setting it at 0 would mask the occasional boxes that encompass nearly the whole image, which is not desirable). This value was set manually by verifying for the intended effect on a few training examples (that is, masking most of the relevant visual evidence).
See \fig\ref{figSuppVqa} for examples of original questions and their counterfactual versions.

\vspace{3pt}\paragraph{Experimental setting.}
Our experiments use a validation set (8,000 questions chosen at random) held out from the original VQA-CP training set. Note that {most existing methods evaluated in VQA-CP use the extremely unsanitary practice of using the VQA-CP test split for model selection}. This is extremely concerning since the whole purpose of VQA-CP is to evaluate generalization to an out-of-distribution test set. The variance in evaluating the `number' and `yes/no' questions is moreover extremely high, because the number of reasonable answers on each of these types is very limited. For example, a model that answers \textit{yes} or \textit{no} at random, or produces constantly either answer, can fare extremely well (upwards of 62\% accuracy) on these questions. This can very well result from a buggy implementation or a ``lucky'' random seed, identified by model selection on the test set (!). This is the reason why we include an evaluation on the `other' type of questions in isolation. All of these issues have been pointed out by a few authors~\cite{grand2019adversarial,teney2019incorporating,teney2019actively}.

Our \textit{focused} test set is a subset of the official VQA-CP test set. It is created in a similar manner as the counterfactual examples. We mask features that overlap with human attention maps \emph{below} (instead of above) a threshold of 0.8. This value was set manually by verifying for the intended effect on a few examples (masking the background but not the regions necessary to answer the question). The \textit{focused} test set is much smaller than the official test set since it only comprises questions for which a human attention map is available.

\vspace{3pt}\paragraph{Models.}
Our baseline model follows the general description of Teney \etal~\cite{teney2017challenge}. We use the features of size 36$\times$2048 provided by Anderson \etal~\cite{anderson2017features}. Our `strong baseline' uses the additional procedure described in~\cite{teney2020unshuffling} on top of this baseline, using the code provided by the authors.


\vspace{3pt}\paragraph{Existing methods.}
The method presented in \cite{selvaraju2019taking} could have constituted an ideal point of comparison with ours, as it was evaluated on VQA-CP and used human attention maps. However, after extensive discussions with the authors, we still have not been able to replicate any of the performance claimed in the paper. We found a number of errors in the paper, as well as inconsistencies in the reported results, and an extreme sensitivity to a single hyperparameter (their reported results were obtained with a single run on a single random seed). We chose not to mention this work in our main paper until these issues have been resolved.

\vspace{3pt}\paragraph{Why not use the same technique for the VQA and COCO experiments ? Inpainting in pixel space \textit{vs} masking image features.}
The two approaches are applicable in both cases. The only reason was to showcase the use of multiple techniques to generate counterfactual examples. The human attention map are specific to VQA and not applicable to the COCO experiments.

\section{Application to image classification with COCO}

\begin{figure}[p]
	\centering
	\renewcommand{\tabcolsep}{0.3em}
	\renewcommand{\arraystretch}{0.8}
	\begin{tabularx}{\linewidth}{cl}
	\hangBoxC{\includegraphics[height=.138\linewidth]{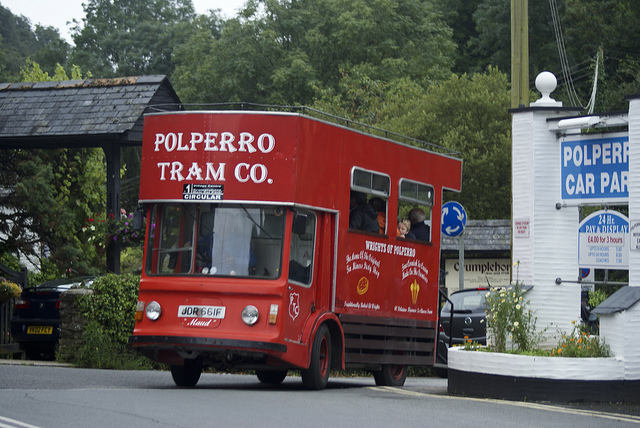}~\includegraphics[height=.138\linewidth]{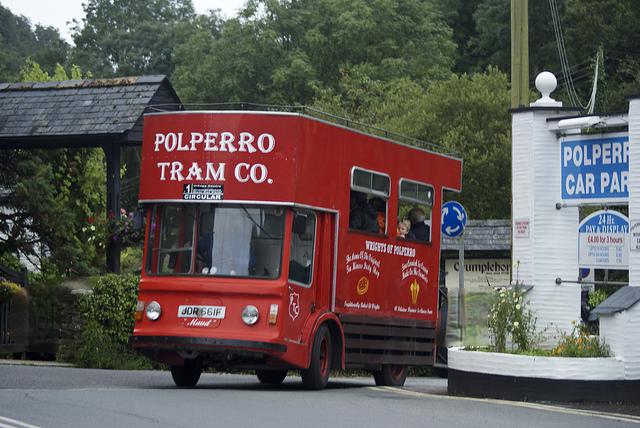}}&\hangBoxC{\footnotesize Masked object: \textbf{car}\\(left and right, behind the truck)}\\
	\hangBoxC{\includegraphics[height=.138\linewidth]{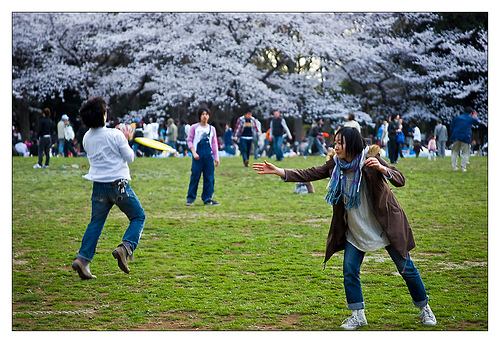}~\includegraphics[height=.138\linewidth]{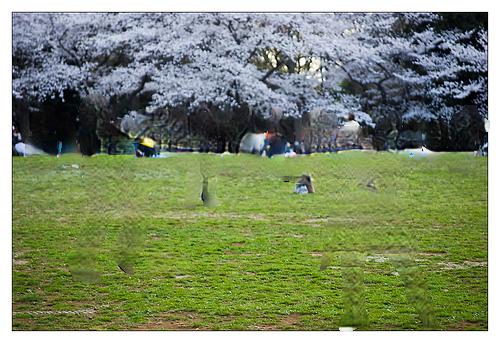}}&\hangBoxC{\footnotesize Masked object: \textbf{person}}\\
	\hangBoxC{\includegraphics[height=.138\linewidth]{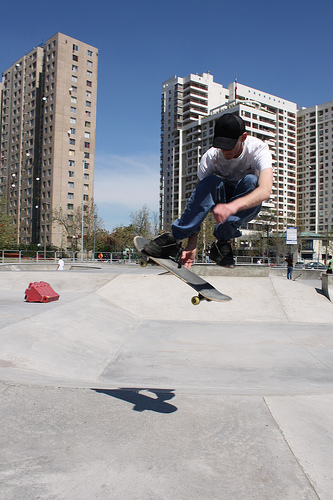}~\includegraphics[height=.138\linewidth]{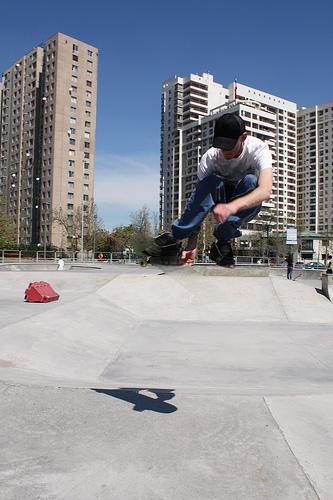}}&\hangBoxC{\footnotesize Masked object: \textbf{skateboard}}\\
	\hangBoxC{\includegraphics[height=.138\linewidth]{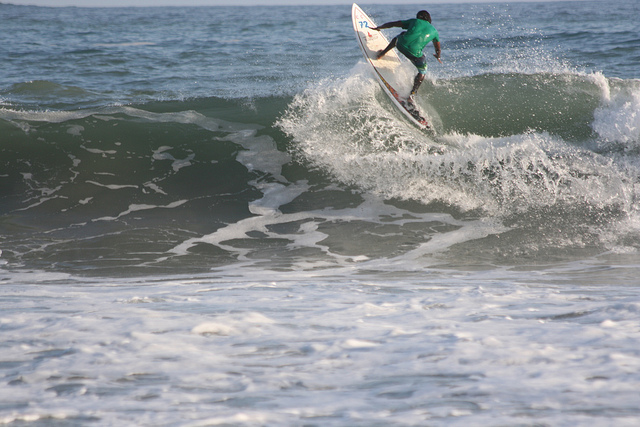}~\includegraphics[height=.138\linewidth]{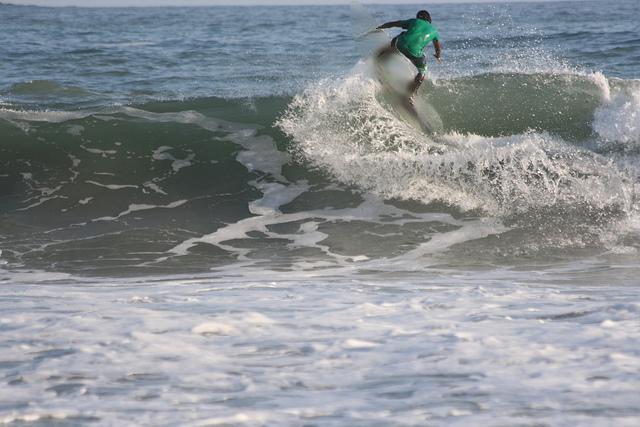}}&\hangBoxC{\footnotesize Masked object: \textbf{surfboard}}\\
	\hangBoxC{\includegraphics[height=.138\linewidth]{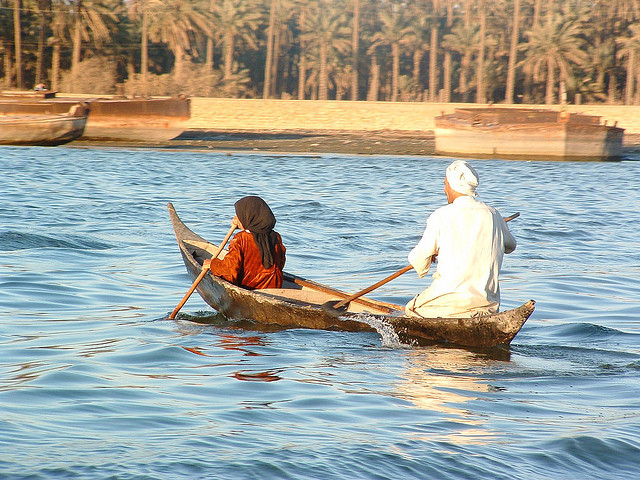}~\includegraphics[height=.138\linewidth]{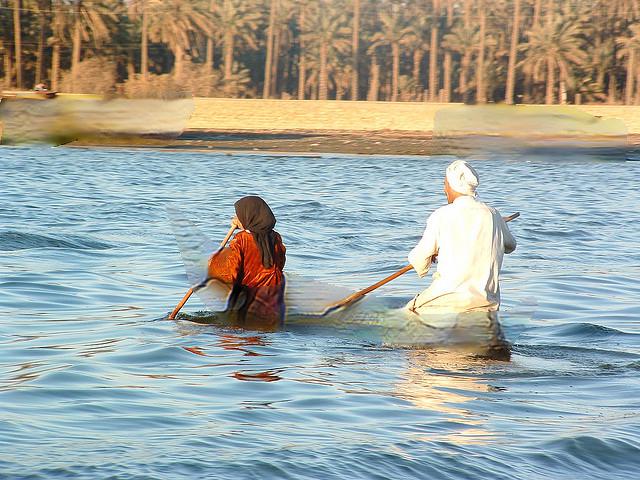}}&\hangBoxC{\footnotesize Masked object: \textbf{boat}}\\
	\hangBoxC{\includegraphics[height=.138\linewidth]{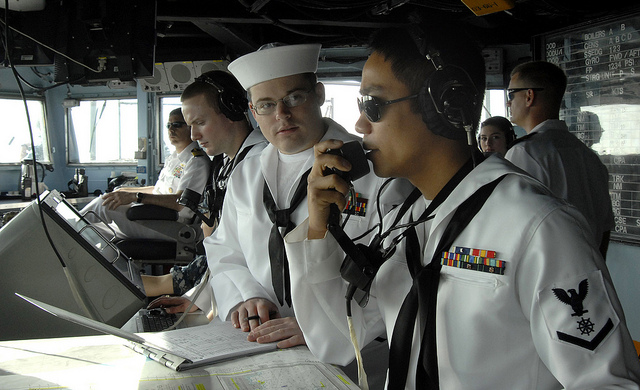}~\includegraphics[height=.138\linewidth]{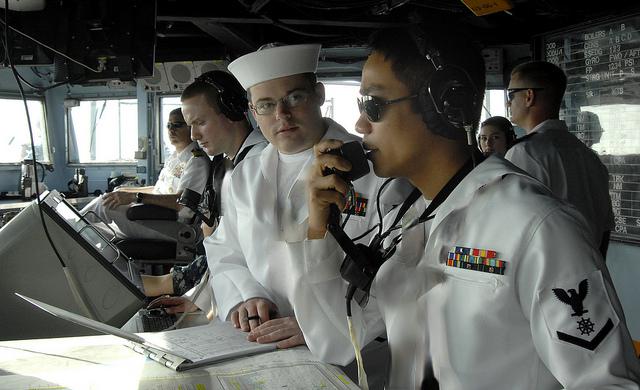}}&\hangBoxC{\footnotesize Masked object: \textbf{tie}\\(on both persons in the foreground)}\\
	\hangBoxC{\includegraphics[height=.138\linewidth]{COCO_val2014_000000003590.jpg}~\includegraphics[height=.138\linewidth]{COCO_val2014_000000003590_000000000002.jpg}}&\hangBoxC{\footnotesize Masked object: \textbf{bicycle}\\(against the railing on the right)}\\
	\hangBoxC{\includegraphics[height=.138\linewidth]{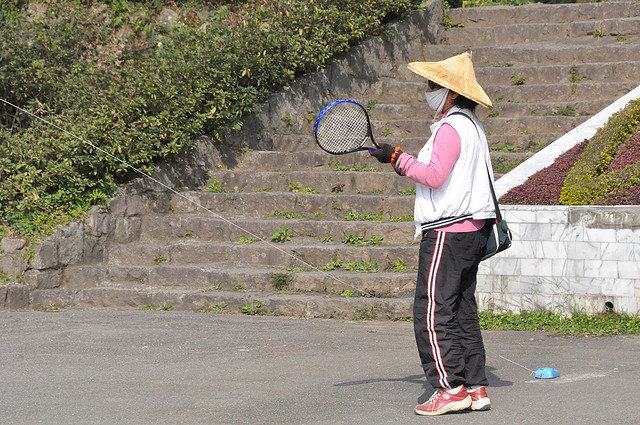}~\includegraphics[height=.138\linewidth]{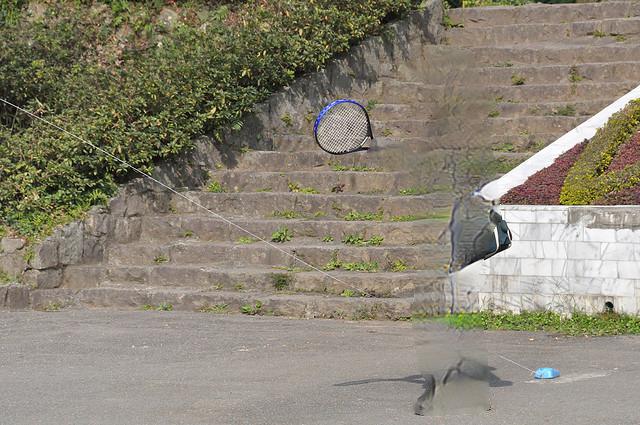}}&\hangBoxC{\footnotesize Masked object: \textbf{person}}\\
	\hangBoxC{\includegraphics[height=.138\linewidth]{COCO_val2014_000000004972.jpg}~\includegraphics[height=.138\linewidth]{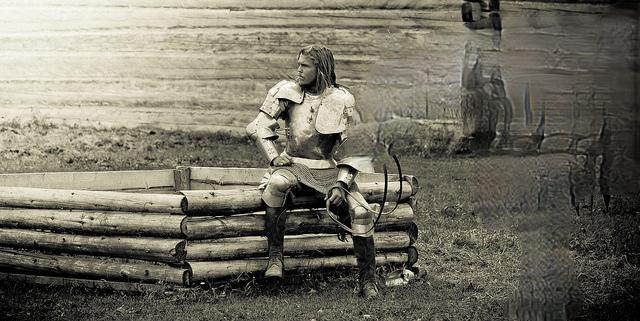}}&\hangBoxC{\footnotesize Masked object: \textbf{horse}}\\
	\hangBoxC{\includegraphics[height=.138\linewidth]{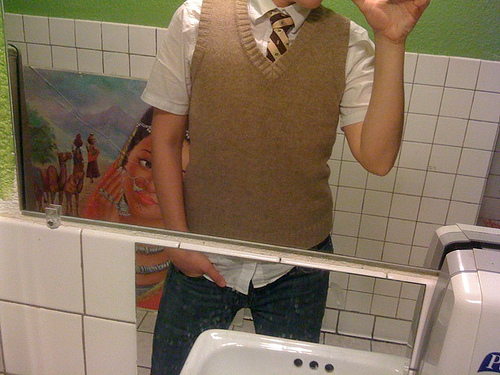}~\includegraphics[height=.138\linewidth]{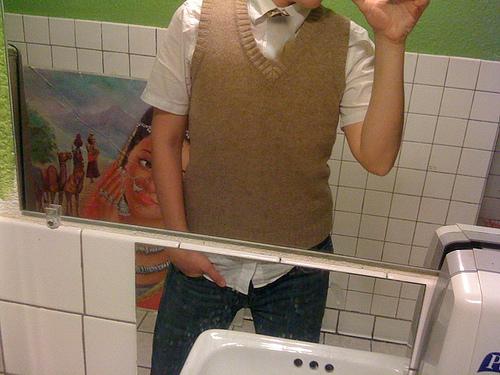}}&\hangBoxC{\footnotesize Masked object: \textbf{tie}}\\
	\end{tabularx}
	\vspace{-7pt}
	\caption{Application to multi-label image classification with COCO. Examples of original and edited images.}
	\vspace{-9pt}
\end{figure}

\vspace{3pt}\paragraph{Data.}
We use the edited images released by~\cite{agarwal2019towards} together with the corresponding original images from COCO.
The edited images were created with the inpainter GAN~\cite{shetty2018adversarial} to mask ground truth bounding boxes of specific objects.
The images come from the COCO splits \textit{train2014} and \textit{val2014}. We keep this separation for our experiments as follows.
Images from \textit{train2014} (323,116~counting original and edited ones) are used for training, except a random subset (1,000 images) that we hold out for validation (model selection, early stopping).
Images from \textit{val2014} (3,361 original and 3,361 edited) are used exclusively for testing.

We identified a subset (named \textit{Hard edited}) of the edited images from \textit{val2014} whose ground truth vector (which indicated the classes appearing in the image) is never seen during training (614 images).

The set of edited images provided by~\cite{agarwal2019towards} is a non-standard subset of COCO, so no directly-comparable results have been published for the multi-label classification task that we consider.

\vspace{3pt}\paragraph{Model.}
We pre-extract image features from all images with the ResNet-based,  bottom-up attention model~\cite{anderson2017features}.
These features are averaged across spatial locations, giving a single vector of dimensions 2048 to represent each image.
Our model is a 3-layer ReLU MLP of size 64, followed by a linear/sigmoid output layer of size 80 (corresponding to the 80 COCO classes).
This baseline model was first tuned for best performance on the validation set (tuning the number of a layers and their size, the batch size, and learning rate), before adding the proposed GS loss.
The model is optimized with AdaDelta, mini-batches of size 512, and a binary cross-entropy loss.

Performance is measured with a standard mean average precision (mAP) (as defined in the Pascal VOC challenge) over all 80 classes.

The \fig\ref{figCoco} in the paper shows the input image with the scores of the top-$k$ predicted labels by the baseline and by our method. The $k$ corresponds to the number of ground truth labels of each image.

\vspace{3pt}\paragraph{Random baseline.}
In our ablations, this model is identical to the standard baseline, but it is trained with a randomly shuffled training set. We shuffle the inputs $\{\bx_i\}_i$ and the ground truth labels $\{y_i\}_i$ of all training examples.
The model is thus not getting any relevant training signal from any example. It can only leverage static dataset biases (\ie a class imbalance).

\section{Application to NLP tasks}

\vspace{3pt}\paragraph{Sentiment analysis data.}
We use the subset of the IMDb dataset~\cite{maas2011learning} for which Kaushik \etal~\cite{kaushik2019learning} obtained counterfactual examples.
We use their `paired' version of the data, which only contains original examples that do have an edited version.
For \textbf{training}, we use the `train' split of original and edited data~(3414~examples).
For \textbf{validation} (model selection, early stopping), we use the `dev' set of paired examples.
For \textbf{testing}, we use the `test' split, reporting accuracy over the original and edited examples separately.
For testing on other datasets, we use a random subset (2000~examples) of the test sets of Amazon Reviews~\cite{ni-etal-2019-justifying}, Semeval 2017 (Twitter data)~\cite{rosenthal-etal-2017-semeval}, and Yelp reviews~\cite{yelpChallenge} similarly to~\cite{kaushik2019learning}.

\vspace{3pt}\paragraph{Sentiment analysis model.}
We first optimized a simple baseline model on the validation set (tuning the number of a layers, embedding sizes, batch size, and learning rate).
We then added the proposed gradient supervision, tuned its hyperparameters on the validation set (regularizer weight) then reported the performance on the test sets at the epoch of best performance on the validation set.
The sentences are tokenized and trimmed to a maximum of 32 tokens.
The model encodes a sentence as a bag of words, using word embeddings of size 50, averaged to the exact length of each sentence (\ie not including the padding of the shorter sentences).
The vocabulary is limited to the 20,000 most frequent words in the dataset.
The averaged vector is passed to a simple linear classifier with a sigmoid output.
All weights, including word embeddings, are initialized from random values, and optimized with AdaDelta, in mini-batches of size 32, with a binary cross-entropy loss.
The best weight for the GS regularizer was found to be $\lambda$=20.
To reduce the noise in the evaluation due to the small size of the training set, we use an ensemble of 6 identical models trained in parallel. The reported results uses the output of the ensemble, that is the average of the logits of the 6 models.

\vspace{3pt}\paragraph{NLI data.}
The experiments on NLI follow a similar procedure to those on sentiment analysis.
We use the subset of the SNLI dataset~\cite{bowman2015large} for which Kaushik \etal~\cite{kaushik2019learning} collected counterfactual examples.
We use their biggest version of the data, named `all combined', that contains counterfactual examples with edited premises and edited hypotheses.
For testing, we evaluate accuracy separately on original and edited examples (edited premises and edited hypotheses combined).
For testing transfer, we use the `dev' set of MultiNLI~\cite{williams2017broad}.
Whereas the SNLI dataset contains sentence pairs derived from image captions, MultiNLI is more diverse. It contains sentences from transcribed speech, popular fiction, and government reports. Compared to SNLI, it contains more linguistic diversity and complexity.

\vspace{3pt}\paragraph{NLI model.}
The premise and hypothesis sentences are tokenized and trimmed to a maximum of 32 tokens. They are encoded separately as bags of words, using frozen Glove embeddings (dimension 300), then a learned linear/ReLU projection to dimension 50, and an average to the length of each sentence (without using the padding). They are then passed through a batch normalization layer, then concatenated, giving a vector of size 100. The vector is passed through 3 linear/ReLU layers, then a final linear/sigmoid output layer. The model is trained with AdaDelta, with mini-batches of size 512, and a binary cross-entropy loss.
The best weight for the GS regularizer was found to be $\lambda$=0.01.
Similarly to our experiments on sentiment analysis, we evaluate an ensemble of 6 copies of the model described above.

\begin{table}[t!]
\scriptsize
\renewcommand{\tabcolsep}{0.15em}
\renewcommand{\arraystretch}{1.01}
\centering
\renewcommand{\tablefont}{\footnotesize}
	\footnotesize
	\begin{tabular}{lcc}
	\Xhline{1\arrayrulewidth}
	Test data	$\rightarrow$	~ & \hspace{4em} & Yelp \\
	\Xhline{1\arrayrulewidth}
	Random predictions (chance) && 45.4 \\
	Baseline w/o edited tr. data && 82.8\\
	Baseline w/ edited tr. data && 87.4 \\
	+ \textbf{GS}, counterfactual rel. && \textbf{88.8} \\
	+ GS, random relations && 57.4 \\ 
	\Xhline{1\arrayrulewidth}
	\end{tabular}
	\normalsize
	\vspace{4pt}
	\normalsize
	\caption{Application to sentiment analysis. Results on the Yelp dataset. This column was missing in \tab\ref{tabSentiment} in the paper (a code-generating error replicated the values from the \textit{Amazon} column into the \textit{Yelp} column).}
	\vspace{-10pt}
\end{table}




\bibliography{Bibliography}

\end{document}